\documentclass{article}

\PassOptionsToPackage{numbers,sort}{natbib}

\usepackage[utf8]{inputenc}
\usepackage{microtype}
\usepackage{graphicx}
\usepackage{subfigure}
\usepackage{amsmath}
\usepackage{amsfonts}
\usepackage{dsfont}
\usepackage{stmaryrd}
\usepackage{booktabs} % for professional tables
\usepackage[hidelinks]{hyperref}
\usepackage[toc,page,header]{appendix}
\usepackage{titletoc}
\usepackage{pdfpages}
\usepackage{enumitem}

\newcommand{\un}{\mathds{1}}

\def\rset{\mathbb{R}}

\def\nset{\mathbb{N}}

\DeclareMathOperator*{\argmax}{argmax}

\newtheorem{remark}{{\bf Remark}}

\newcommand{\expec}{\mathbb{E}}
\usepackage{tikz}
\usetikzlibrary{shapes.geometric, arrows}
\tikzstyle{block} = [rectangle, rounded corners, minimum width=0.5cm, minimum height=0.6cm,align=center,text centered, draw=black]
\tikzstyle{arrow} = [thick,->,>=stealth]
\usepackage[final]{neurips_2022}
\title{MARLIM: Multi-Agent Reinforcement Learning \\ for Inventory Management}

\author{%
	R\'{e}mi Leluc\thanks{\href{mailto:remi.leluc@gmail.com}{remi.leluc@gmail.com}, Work done while being at TotalEnergies OneTech} \\
	CMAP, \'{E}cole Polytechnique \\
	TotalEnergies OneTech 
	\And
	Elie Kadoche\thanks{\{first.last\}@totalenergies.com, TotalEnergies OneTech, 2 Place Jean Millier, 92400 Courbevoie, France} \\
	TotalEnergies OneTech 
	\AND
	Antoine Bertoncello\footnotemark[2] \\
	TotalEnergies OneTech 
	\And
	S\'{e}bastien Gourv\'{e}nec\footnotemark[2] \\
	TotalEnergies OneTech 
}

\begin{document}




%\begin{icmlauthorlist}
%\icmlauthor{Rémi Leluc}{tp,total}
%\icmlauthor{Elie Kadoche}{total}
%\icmlauthor{Antoine Bertoncello}{total}
%\icmlauthor{Sébastien Gourvénec}{total}
%\end{icmlauthorlist}
%\icmlaffiliation{total}{TotalEnergies OneTech (3 Boulevard Thomas %Gobert, 91120 Palaiseau, France). E-mail addresses: %\{firstname.lastname\}@totalenergies.com}
%\icmlaffiliation{tp}{Télécom Paris, IPParis (19 place Marguerite %Perey, 91120 Palaiseau, France);}
%\icmlcorrespondingauthor{Rémi Leluc}{remi.leluc@gmail.com} 

\maketitle

\begin{abstract} Maintaining a balance between the supply and demand of products by optimizing replenishment decisions is one of the most important challenges in the supply chain industry. This paper presents a novel reinforcement learning framework called MARLIM, to address the inventory management problem for a single-echelon multi-products supply chain with stochastic demands and lead-times. Within this context, controllers are developed through single or multiple agents in a cooperative setting. Numerical experiments on real data demonstrate the benefits of reinforcement learning methods over traditional baselines.
\end{abstract}

\section{Introduction}
\label{sec:intro}

Inventory control  \citep{silver1985decision} is one of the major problems in the supply chain industry. The main goal is to ensure the right balance between the supply and demand of products by optimizing replenishment decisions. More precisely, a controller observes the past demands and local information of the inventory and has to decide about the next ordering values. Accurate inventory management is key to running a successful product business with benefits ranging from better inventory accuracy and insights to cost savings and avoidance of shortage and stock overflows. 

The main issue of the supply chain is the environment uncertainty \citep{zipkin2000foundations}. In an ideal world with deterministic demand and lead-times, an inventory controller would be able to place a perfect order equal to the demand size at the right time. However, in practice, both the demands and lead-times are stochastic with potentially high volatility, making the inventory management problem hard to solve. In most cases, the inventory controller may exceedingly or insufficiently order. The former case leads to paying unnecessary ordering and holding costs while the latter results in shortage costs which may jeopardize the company’s performance.

Classical methods for solving the inventory management problem rely on basic heuristics \citep{zheng1991finding,toomey2000inventory} due to the complexity of the mathematical modeling of the inventory system. While these approaches are easy to implement, they are not able to capture the randomness of the demand and lead-times. Another way of solving the inventory management problem is dynamic programming \citep{bellman1966dynamic}. Despite being efficient, this technique requires exact knowledge of the mathematical model of the inventory system, which becomes intractable for big companies with very large inventories. Inventory management models can quickly become too complex and time-consuming, leading to unworkable models \citep{gijsbrechts2021can}. To escape this curse of dimensionality, one may apply approximate dynamic programming \citep{halman2009fully,fang2013sourcing} which performs well in specific settings at the cost of strong assumptions and simplifications.

Since the environment uncertainty is the major problem of inventory control \citep{chaudhary2018state}, reinforcement learning (RL) methods appear as a natural solution since they can model complex situations and generalize well in a data-driven manner. The reinforcement learning task \citep{sutton2018reinforcement} consists, for an autonomous agent (e.g., a robot), in learning the actions to be taken, from trials, in order to optimize a quantitative signal over time. Such paradigm has reached tremendous success in games \citep{mnih2015human,silver2018general,vinyals2019grandmaster}, but the early applications of reinforcement learning to real-world tasks, e.g., robotics \citep{kober2013reinforcement} or autonomous driving \citep{okuda2014survey,sallab2017deep} remain a challenge and an active field of research. Furthermore, most of the literature on inventory management systems is meant to be applied for companies which are specialized in the retail industry where \textit{(i)} the items of the inventory are meant to be sold and \textit{(ii)} item shortages may be addressed through back orders. However, many companies are interested in inventory management for factories and warehouses where the objective is to ensure the performance of a production line and avoid the drastic consequences of item shortages.

The main goal of this paper is to develop a novel reinforcement learning framework, called MARLIM, to address the inventory management problem for a single-echelon multi-products supply chain on a production line with stochastic demands and lead-times.

\textbf{Related work.}
%\subsection{Related work}
The stochastic inventory control problem is one of the most studied problems in inventory theory. It was initiated by the seminal works of \cite{fukuda1964optimal} and \cite{silver1985decision} which presented the basic heuristics to solve the inventory management problem. The so-called (s,S) policy \citep{zheng1991finding}, which results from the economies of scale in procurement, is a widely used policy nowadays in practice. However such heuristic cannot capture the randomness of the demand and lead-times, making it difficult to apply to realistic use cases. Later on, several authors developed the theory of inventory management and one may refer to the books of \cite{zipkin2000foundations,porteus2002foundations} and \cite{levi2014logic} for an in-depth coverage of the topic. The different state-of-the-art models for inventory management systems may be found in the recent survey of \cite{chaudhary2018state}.

In the past few years, many comprehensive studies related to the integration of reinforcement learning techniques to the inventory management problem were presented, starting with the work of \cite{giannoccaro2002inventory} which considered production and distribution functions of global supply chain with multiple stages assuming a single item. Later on, different variations were derived with a particular focus on a supply chain with two echelons \citep{kim2008asynchronous,kwon2008case}, multiple echelons \citep{kim2005adaptive,kwak2009situation,chaharsooghi2008reinforcement,sun2012analyses} or multiple retailers \citep{jiang2009case,dogan2015reinforcement}. In all these studies, reinforcement learning methods have been implemented to specify near-optimal ordering policies in the entire supply chain with different goals such as maximizing the profit (when considering retailers) or minimizing costs composed of either ordering and holding costs or holding and backorder costs but not all of them at the same time. Furthermore, all the mentioned studies are dealing with a simplified assumption of independent products with no interactions among them.

In order to handle multiple items and accurately model the inter-dependency between them, the proposed framework of this paper is based on the multi-agent reinforcement learning paradigm \citep{nguyen2020deep} where each item may be seen as a single agent. Such framework seems promising to improve the training efficiency as suggested by the recent survey of \cite{zhang2021multi} and the two recent studies of multi-agent reinforcement learning applied to inventory management problems \citep{barat2019actor,sultana2020reinforcement}. In \cite{barat2019actor}, the authors derive a reinforcement strategy for a grocery retailer with multiple items and fixed delivery times where the focus is on the items availability rather than their associated costs. Similarly the study of \cite{sultana2020reinforcement} deals with a multiple echelons supply chain with predefined lead times whose aim is to maximise product sales and minimise wastage of perishable products. Conversely, the developed method of this paper is focused on the reduction of operating costs on a production line with stochastic demands and lead-times for real-world applications.

The framework of MARLIM is also related to the joint replenishment problem, i.e., when one considers the interdependency among different groups of products in a same order provided by a single supplier \citep{khouja2008review,wang2012differential,salameh2014joint}. The objective is to optimize a global replenishment cost composed of inventory ordering and holding costs \citep{qu2015contrastive,wang2015improved} but does not take into account the losses induced by item shortages. In contrast to these previous studies, the aim of the proposed method whose is not only to reduce the global replenishment cost but also to avoid stock-outs of items. 

\newpage
\textbf{Contributions.}
%\subsection{Contributions}
The main contributions of this paper may be summarized as follows. \\
\textbullet \ A novel reinforcement learning framework, called MARLIM, is developed to address the inventory management problem for a single-echelon multi-products supply chain on a production line with stochastic demands and lead-times. \\
\textbullet \ A methodology to train agents in different scenarios for fixed or shared capacity constraints with specific handling of storage overflows is provided. \\
\textbullet \  Various numerical experiments on real-world data demonstrate the benefits of the developed method over classical baselines.

\textbf{Outline.}
%\subsection{Outline}
Section \ref{sec:preliminaries} presents the mathematical background of reinforcement learning and Section \ref{sec:IM_model} deals with the inventory management model. The methodology and details of the developed supply chain environment are described in Section \ref{sec:marlim}. Numerical experiments are performed in Section \ref{sec:simus} to highlight the relevance of the developed model on real-world data and Section \ref{sec:conclusion} concludes the article with further discussion.

%\textbf{Notation.} For all $t \geq 1$, any feature $u_t=(u_t^{(1)},\ldots,u_t^{(n)}) \in \rset^n$ represents a particular attribute of the whole inventory management system at time $t$ and for all $i=1,\ldots,n$, $u_t^{(i)}$ is the signal associated to the $i^{th}$ product.

\section{Preliminaries on Reinforcement Learning}

\label{sec:preliminaries}

%\subsection{Reinforcement Learning framework}

\textbf{Markovian setting.} %In order to capture the main aspects -sensation, action and goal- of the reinforcement learning problem
%One can define the interaction between a learning agent and its environment in terms of states, actions, and rewards. 
Markov Decision Processes (MDP) \cite{puterman1994markov} are a formalization of sequential decision making, where actions influence not just immediate rewards, but also subsequent situations. Consider the classical framework of a MDP defined as a tuple $\mathcal{M}=\langle\mathcal{S}, \mathcal{A}, p, r, \gamma, \mu\rangle,$ comprised of a state space $\mathcal{S},$ an action space $\mathcal{A},$ a Markovian transition kernel $ p : \mathcal{S} \times \mathcal{A} \rightarrow \Delta(\mathcal{S})$ where $\Delta(\mathcal{S})$ denotes the set of probability density functions over $\mathcal{S},$ a reward function $r : \mathcal{S} \times \mathcal{A} \rightarrow \mathbb{R},$ a discount factor $\gamma \in(0,1)$ and an initial-state distribution $\mu \in \Delta(\mathcal{S})$. The reinforcement learning problem consists of finding the best strategy or policy $\pi: \mathcal{S} \to \mathcal{A}$ in order to maximize the performance. The criterion here is defined in terms of future rewards that can be expected and depend on the agent's behavior.
The solution of a MDP is an optimal policy $\pi^{*}$ that maximizes the value function $V^\pi$ %, or equivalently the state-action value function $Q^\pi$ 
in all the states $s \in \mathcal{S}$ over some policy set of interest $\Pi$: $\pi^{\star} \in \argmax_{\pi \in \Pi} V^{\pi}$. Such optimal policy $\pi^\star$ is guaranteed to exist thanks to the theorem of \cite{bellman1959functional}. In practice, $\pi^\star$ can be found through dynamic programming and Bellman equations \cite{bellman1957dynamic,bertsekas1996neuro,howard1960dynamic} with different schemes, e.g., policy iteration \cite{bertsekas2011approximate,bucsoniu2012least} or value iteration \cite{pineau2003point,sondik1971optimal}. Another way is to consider policy-gradient methods \citep{williams1992simple,baxter2001infinite} which use a parameterized policy $\pi_{\theta}$ with $\theta \in \Theta \subset \rset^d$ and update the policy parameter $\theta$ on each step in the direction of an estimate of the gradient of the performance with respect to the policy parameter.

%\begin{definition}(Bellman equations) The value functions are related to the values at different states through the Bellman equations. For the discounted formulation, we have
%\begin{align*}
%&V^{\pi}(s) = \sum_{a \in \mathcal{A}} \pi(a|s) \sum_{s' \in \mathcal{S}}\left(r(s,a) + \gamma V^{\pi}(s')\right)  p(s'|s,a)   , \\
%&Q^{\pi}(s,a) = r(s,a)+  \gamma \sum_{s' \in \mathcal{S}} p(s'|s,a) V^{\pi}(s').
%\end{align*}
%\end{definition}

\textbf{Parametric policies.}
%%%%%%%%%%%%%%%%%%%%%%%%%%%%%%%%
Given a parameter space $\Theta \subseteq \rset^d$, consider a class of smoothly parameterized stochastic policies $\Pi_{\Theta}=\left\{\pi_{\theta}, \theta \in \Theta\right\}$ that are twice differentiable w.r.t. $\theta$, \textit{i.e.}, for which the gradient $\nabla_{\theta} \pi$ and the Hessian $\nabla_{\theta}^{2} \pi_{\theta}$ are defined everywhere and finite. When the action space is finite, a popular choice is to use \textit{Gibbs policies}, a.k.a. \textit{softmax policies}, for all $s \in \mathcal{S}$ and $a \in \mathcal{A}$,
\begin{align} \label{eq:gibbs}
\pi_{\theta}(a|s)=\frac{\exp(\theta^T \psi(s,a))}{\sum_{a^\prime \in \mathcal{A}} \exp(\theta^T \psi(s,a^\prime))}
\end{align}
where $\psi: \mathcal{S} \times \mathcal{A} \rightarrow \rset^d$ is an appropriate feature-extraction function, often computed using a neural network. When the action space is continuous $\mathcal{S} \subset \rset^d$, a popular choice is to use \textit{Gaussian policies} so the policy can be defined as the normal probability density over a real-valued scalar action,  with some parametric mean $\mu: \Theta \times \mathcal{S} \rightarrow \rset$ and standard deviation $\sigma: \Theta \times \mathcal{S} \rightarrow \rset_{+}$ that depend on the state,  for all $s \in \mathcal{S}$ and $a \in \mathcal{A}$,
\begin{align} \label{eq:gaussian}
\pi_{\theta}(a | s)= \frac{1}{\sigma_{\theta}(s) \sqrt{2 \pi}} \exp \left(-\frac{(a-\mu_{\theta}(s))^{2}}{2 \sigma_{\theta}(s)^{2}}\right).
\end{align}
%More recently, \cite{chou2017improving} suggested to consider \textit{Beta} distributions in order to avoid bias due to boundary effects when working with bounded continuous action spaces, using parameters $\alpha = \alpha_{\theta(s)}, \beta = \beta_{\theta(s)}$ and the policy
%\begin{align} \label{eq:beta}
%\pi_{\theta}(a | s)= \frac{\Gamma(\alpha + \beta)}{\Gamma(\alpha)\Gamma(\beta)} a^{\alpha-1}(1-a)^{\beta-1}.
%\end{align}

%\subsection{Multi-Agent Reinforcement Learning}
\textbf{Multi-Agent Reinforcement Learning (MARL).} Originated from the seminal works of \cite{shapley1953stochastic} and \cite{littman1994markov}, Markov Games are the standard generalization of Markov Decision Processes as they capture the interaction of multiple agents. A Markov game is defined as a tuple $\langle \mathcal{N}, \mathcal{S},(\mathcal{A}^i)_{i \in \mathcal{N}},p,(r^i)_{i \in \mathcal{N}},\gamma \rangle$ where $\mathcal{N}$ denotes the set of agents, $\mathcal{S}$ is the state space observed by all agents, $\mathcal{A}^i$ is the action space of agent $i$. $\mathcal{A}= \prod_{i \in \mathcal{N}} \mathcal{A}^i$ denotes the joint action space, $p:\mathcal{S} \times \mathcal{A} \to \Delta(\mathcal{S})$ is the transition kernel, $r^i : \mathcal{S} \times \mathcal{A} \rightarrow \mathbb{R}$ is the immediate reward function of agent $i$ and  $\gamma \in(0,1)$ is the discount factor.
%\begin{definition}(Markov game)
%A Markov game is defined as a tuple $\langle \mathcal{N}, \mathcal{S},(\mathcal{A}^i)_{i \in \mathcal{N}},p,(r^i)_{i \in \mathcal{N}},\gamma \rangle$ where $\mathcal{N}$ denotes the set of agents, $\mathcal{S}$ is the state space observed by all agents, $\mathcal{A}^i$ is the action space of agent $i$. $\mathcal{A}= \prod_{i \in \mathcal{N}} \mathcal{A}^i$ denotes the joint action space, $p:\mathcal{S} \times \mathcal{A} \to \Delta(\mathcal{S})$ is the transition kernel, $r^i : \mathcal{S} \times \mathcal{A} \rightarrow \mathbb{R}$ is the immediate reward function of agent $i$ and  $\gamma \in(0,1)$ is the discount factor.
%\end{definition}
At time step $t$, each agent $i \in \mathcal{N}$ selects an action $a_t^{(i)}$ based on the system state $s_t$. The system then transitions to state $s_{t+1}$ according to $p$ and rewards each agent $i$ with $r^i(s_t,a_t)$. The goal of agent $i$ is to optimize its own long-term reward by finding the policy $\pi^{(i)}:\mathcal{S} \to \Delta(\mathcal{A}^{i})$ such that $a_t^{(i)} \sim \pi^{(i)}(\cdot|s_t)$. Therefore, the marginal value function $V^{(i)}: \mathcal{S} \to \rset$ of agent $i$ becomes a function of the joint policy $\pi:\mathcal{S} \to \mathcal{A}$ defined by $\pi(a|s) = \prod_{i \in \mathcal{N}} \pi^{(i)}(a^{(i)}|s), V^{(i)}(s) = \expec\left[\sum_{t \geq 0} \gamma^t r^{i}(s_t,a_t) \Big \rvert a_t^{(i)} \sim \pi^{(i)}(\cdot|s_t), s_0=s \right].$ Thus, the solution concept of a Markov game deviates from that of a MDP since the optimal performance of each agent is controlled not only by its own policy, but also the choices of all other players of the game. The most common solution concept for Markov games is Nash equilibrium \citep{bacsar1998dynamic,filar2012competitive}. %Nash equilibrium characterizes an equilibrium point $\pi_{\star}$, from which none of the agents has any incentive to deviate. In other words, for any agent $i \in \mathcal{N}$, the policy $\pi_{\star}^{i}$ is the best response to $\pi_{\star}^{-i}$, where $-i = \mathcal{N}\backslash\{i\}$ represents the set of all agents in $\mathcal{N}$ excepts agent $i$.

%\begin{definition}(Nash Equilibrium) A Nash equilibrium of the Markov game $\langle \mathcal{N}, \mathcal{S},(\mathcal{A}^i)_{i \in \mathcal{N}},p,(r^i)_{i \in \mathcal{N}},\gamma \rangle$ is a joint policy $\pi_{\star}$ such that for any $s \in \mathcal{S}$ and $i \in \mathcal{N}, V_{\pi_{\star}}^{(i)}(s) \geq V_{\pi^i,\pi_{\star}^{-i}}^{(i)}(s)$ for any $\pi^i$.
%\end{definition}
As a standard learning goal for MARL, Nash equilibrium always exists for finite-space infinite-horizon discounted Markov games \citep{filar2012competitive}, but may not be unique in general. Most of the multi-agent reinforcement learning algorithms are contrived to converge to such an equilibrium point, if it exists.
\section{Inventory Management Model}
\label{sec:IM_model}

Denote by $\mathcal{N}=\{1,\ldots,n\}$ the discrete product space where $i \in \mathcal{N}$ refers to the $i^{th}$ item in the inventory and $n$ is the total number of items. At each time step, the inventory controller decides about the order to take based on the current inventory level and the previous demands. This order arrives in the inventory after a stochastic lead-time. After receiving the replenishment quantity of different products, the demands are satisfied and the environment incurs inventory costs. The aim of this Section is to describe the complete inventory management model from the inventory features and inventory costs to the inventory dynamics of the system.

\subsection{Inventory features} \label{subsec:features}
\begin{minipage}{0.45\textwidth}
The different inventory features are the basis of the inventory management model as they model the state space and the dynamics of the underlying Markov game. %Similarly, the inventory costs are the key to represent the rewards of the differents agents.
 The inventory level $x_t \in \rset^n$ is the on-hand inventory at time $t$, i.e., the quantity of products sitting on the shelf in the inventory. At each time step, the inventory controller receives a stochastic demand signal $\delta_t \in \rset^n$ and can order a quantity $a_t \in \rset^n$ of items. These quantities arrive in the inventory after some lead-time $\tau_t \in \rset^n$. The inventory replenishment quantity $\rho_t \in \rset^n$ is the associated quantity of products which is on its way to the inventory, i.e., it is the quantity that will be added to the inventory level when the inventory check is done at the next time step.
 \end{minipage}
\hfill
\begin{minipage}{0.56\textwidth}
% \begin{minipage}[t]{0.54\textwidth}
%\flushright
\centering
    \begin{tabular}{|c|l|}
    \hline
       \text{Symbol} & \multicolumn{1}{c|}{\text{Feature}} \\
    \hline
        $\Lambda^{(k)}$ & \text{maximum capacity of subspace $k$} \\
        $x_t^{(i)}$ & \text{inventory level of product $i$ at time $t$} \\
        $a_t^{(i)}$ & \text{order of product $i$ at time $t$} \\
        $\rho_t^{(i)}$ & \text{replenishment  of product $i$ at time $t$} \\
        $\lambda_t^{(i)}$ & \text{temporary level of product $i$ at time $t$} \\
        $w_t^{(i)}$ & \text{overflow weight of product $i$ at time $t$} \\
        $\Omega_{t,k}^{\phantom{()}}$ & stock overflow of subspace $k$ at time $t$ \\
        
        $\delta_t^{(i)}$ & \text{demand of product $i$ at time $t$} \\
        $\tau_t^{(i)}$ & \text{lead-time of product $i$ at time $t$} \\
        $\beta_t^{(i)}$ & \text{backlog of product $i$ at time $t$} \\
        \hline
        $C_o^{(i)}$ & \text{Ordering cost of product $i$ } \\
        $C_h^{(i)}$ & \text{Holding \ cost of product $i$ } \\
        $C_s^{(i)}$ & \text{Shortage cost of product $i$ } \\
        \hline
        \multicolumn{2}{c}{}
        %Inventory features.
\end{tabular}
%\caption{Inventory features.}
\label{tab:features}
\end{minipage}
When receiving the replenishment quantity, the inventory levels are temporarily updated through $\lambda_t \in \rset^n$ and some storage overflow may happen. When this overflow event happens, denoted by $\Omega_t$, the replenishment quantities are scaled using weights $w_t \in \rset^n$ to ensure that the total inventory level does not exceed the maximum storage capacity $\Lambda \in \nset^{+}$. Finally, the demands are satisfied and the controller updates the inventory levels $x_{t+1}$. When demand exceeds the available inventory level for an item, it yields a shortage cost computed through a backlog level $\beta_{t+1} \in \rset^n$. When optimizing the decisions relative to the inventory, the controller must take into account the different inventory costs.
\subsection{Inventory costs} \label{subsec:costs} The inventory costs may be classified into three categories \citep{toomey2000inventory}: ordering, holding and shortage.

% \begin{figure}[tph]
(i) \textit{Ordering costs}: in the case of an ordering setting, these costs include the functioning costs, reception and tests costs, the salaries of the personnel, information systems costs and customs costs. In the case of a production setting, these costs include the raw material costs, labor costs, fixed and variable overheads, e.g., rent of a factory or the energy consumption allocated for production. \\

(ii) \textit{Holding costs}: these are associated with storing inventory that remains unused and may be divided into two categories: financial and functional costs. The former represent the financial interest of the money invested in procuring the stocked products. The latter include the rent and maintenance of the required space, insurance costs, equipment costs, inter-warehouses transportation and obsolescence costs.
% \end{figure}

(iii) \textit{Shortage costs}: when demand exceeds the available inventory for an item. The related costs fall in one of the two following categories: lost sales costs and backlogging costs. When considering lost sales, the unsatisfied demands are completely lost whereas with backlogging costs, there is a penalty shortage cost. Note that in the case where the stock is internal, the inventory shortage will induce the stop of production of therefore all the consequent costs.

For any item $i \in \mathcal{N}$, denote by $C_o^{(i)},C_h^{(i)}$ and $C_s^{(i)}$ the unit ordering, holding and shortage costs respectively. The different features of the inventory management system are summarized in the Table above.

\begin{remark}\label{rem:costs_constant} (Costs and priority) The different unit costs are the keystone to accurately model different behaviors among the items. Intuitively, items with high shortage cost correspond to critical items that should not run out-of-stock. On the contrary, items with high ordering cost should be ordered sparingly. In general, an optimal controller should consider some reward function which leverages these three factors to find the right balance between the different costs.
\end{remark}

\subsection{Inventory Dynamics} \label{subsec:dynamics} The different relations between the inventory features are the mainstay to model a Markov game. First of all, there is a capacity constraint on the storage space and several structures are to be considered. On the one hand, a natural assumption is to consider that each item $i \in \mathcal{N}$ has a maximum capacity $\Lambda^{(i)} \in \nset^\star$. In practice, thanks to expert's knowledge, an estimate of the mean demand of each item may be available so that the capacity of each product can be upper bounded. On the other hand, for economical reasons, the total storage space may be shared among all items. This model allows more flexibility as products can compete for storage space but it assumes that all items belong to a same category. A unified and more realistic setting is the following: the different products may be classified into different types and they only compete for storage space inside their category (see Remark \ref{rem:clusters}). In other words, the product space $\mathcal{N}$ can be decomposed into $K$ different clusters of products, i.e.,  
\begin{align*}
\mathcal{N} = \bigcup_{k=1}^K \mathcal{N}_k.
\end{align*}
Each product subspace $\mathcal{N}_k$ has a maximum storage capacity $\Lambda^{(k)} \in \nset^\star$. For the sake of clarity, all items are assumed to have the same volume (see Remark \ref{rem:capacity}). At each time step $t \geq 0$, the available storage space of each product subspace should be non-negative, i.e., for all $k=1,\ldots,K$
\begin{align}
    \label{eq:capacity}
    \Delta_t^{(k)} = \Lambda^{(k)} - \left(\sum_{i \in \mathcal{N}_k} x_{t}^{(i)} \right) \geq 0.
\end{align}
 The replenishment quantity $\rho_t$ depends on the previous orders $a_1,\ldots,a_t$ and their associated lead-times. Indeed, the received quantity of product $i$ at time $t$ corresponds to all the previous orders $a_j^{(i)}$ made at time step $j$ such that the receiving time $(j+\tau_j)$ obtained by adding the associated lead-time matches the current time step $t$, i.e., for $i=1,\ldots,n$,
 \begin{align*}
    \rho_{t}^{(i)} &= \sum_{j=1}^t a_{j}^{(i)} \un_{\{j+\tau_{j}^{(i)}=t\}}.
 \end{align*}
When receiving these replenishment quantities at time $t$, the inventory levels are temporarily updated as
%may happen that the associated storage spaces overflow. In that case, the new inventory level is set to the maximum capacity of the corresponding item, i.e., for $i=1,\ldots,n$, 
%\begin{align*}
    $\lambda_{t}^{(i)} = x_{t}^{(i)} + \rho_{t}^{(i)}$
%\end{align*}
and it may happen that the storage space of subspace $\mathcal{N}_k$ overflows which translates into the following event
\begin{align*}
    \Omega_{t,k} = \left\{ \sum_{i \in \mathcal{N}_k} \lambda_t^{(i)} > \Lambda^{(k)} \right\}.
\end{align*}
In virtue of Eq.\eqref{eq:capacity}, on $\Omega_{t,k}$ it holds $\sum_{i \in \mathcal{N}_k} \rho_t^{(i)} > \Delta_t^{(k)}$. This means that when the stock overflow happens in $\mathcal{N}_k$ then the received replenishment quantities of products in $\mathcal{N}_k$ exceed the available storage space of $\mathcal{N}_k$. In that case, the replenishment quantities are weighted to completely fill the available storage space, i.e., some overflow weights $w_{t}^{(i)}$ are chosen such that $\sum_{i \in \mathcal{N}_k} w_{t}^{(i)} \rho_{t}^{(i)} = \Delta_t^{(k)}$. In order to promote the replenishment of critical products (see Remark \ref{rem:overflow}), the overflow weights are set to be proportional to the shortage cost of items. For all $k=1,\ldots,K$ and $i \in \mathcal{N}_k$,
\begin{align}
    \label{eq:weight}
     w_t^{(i)} &= \un_{\Omega_{t,k}^{\phantom{c}}} \left(\frac{\Delta_t^{(k)} C_s^{(i)}}{\sum_{j \in \mathcal{N}_k} C_s^{(j)}\rho_t^{(j)}}\right) + \un_{\Omega_{t,k}^c}.
\end{align}

\begin{figure}[h]
\begin{minipage}{0.37\textwidth}
Since the quantities of products are non-negative integers, only the integer part of the weighted replenishment quantities $w_{t} \rho_t$ are added to the current inventory levels $x_t$. When the demands exceed the available on-hand inventory levels, a backlog is computed to monitor the unmet demands and measure the associated shortage costs.
\end{minipage}
\hfill
\begin{minipage}[h]{0.62\textwidth}
\flushright
%\begin{figure}
%\centering
\begin{tikzpicture}[node distance=2cm]
% Draw Nodes
\node (start) [block] {Item $i$, time $t$ \\ level $x_t^{(i)}$};
\node (action) [block, right of=start, xshift=1.2cm] {take order \\ action $a_t^{(i)}$};
\node (orders) [block, right of=action, xshift=1.3cm] {update \\ orders list};
\node (replenishment) [block, below of= orders,yshift=0.6cm] {receive \\ quantity $\rho_t^{(i)}$};
\node (weights) [block, left of= replenishment,xshift=-1.3cm] {overflow \\ weights $w_t^{(i)}$};
\node (update) [block, left of= weights,xshift=-1.2cm] {$(x_{t+1}^{(i)},\beta_{t+1}^{(i)})$};
% Draw Arrows
\draw [arrow] (start) -- node[anchor=south] {} (action);
\draw [arrow] (action) -- node[anchor=north] {lead-time} node[anchor=south] {$\tau_t^{(i)}$} (orders);
%\draw [arrow] (action) -- node[anchor=north] {test} (orders);
\draw [arrow] (orders) -- node[anchor=south] {} (replenishment);
\draw [arrow] (replenishment) -- node[anchor=south] {compute} (weights);
\draw [arrow] (weights) -- node[anchor=south] {update} (update);
\draw [arrow] (update) -- node[anchor=west] {$t \xleftarrow{} t+1$} (start);
\end{tikzpicture}
\caption{Inventory Dynamics}
\label{fig:inventory_dynamics}
%\end{figure}
\end{minipage}
\end{figure}
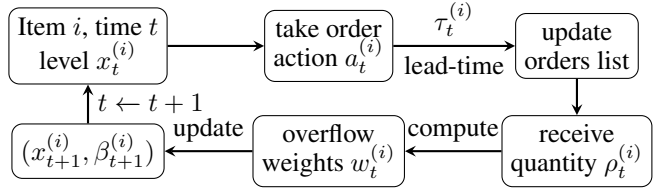

For each item $i \in \mathcal{N}$, the inventory level and backlog are updated as
\begin{align*}
    %\label{eq:update_level}
     x_{t+1}^{(i)} = \left(x_{t}^{(i)} + \lfloor w_{t}^{(i)} \rho_t^{(i)} \rfloor- \delta_t^{(i)}\right)_{+}, \quad 
    \beta_{t+1}^{(i)} = \beta_{t}^{(i)} + \left(x_{t}^{(i)} + \lfloor w_{t}^{(i)} \rho_t^{(i)} \rfloor - \delta_t^{(i)}\right)_{-}
\end{align*}

%Then, the inventory controller receives the demands $\delta_t$ of the different products and the inventory levels are updated accordingly. When the demands exceed the available on-hand inventory levels, a backlog is computed to monitor the unmet demands and measure the associated shortage costs
%\begin{align}
%    \label{eq:update_level}
%     x_{t+1}^{(i)} &= \left(\lambda_{t}^{(i)}- \delta_t^{(i)}\right)_{+} \\
%     \label{eq:backlog}
%    \beta_{t+1}^{(i)} &= \left(\lambda_{t}^{(i)}  - \delta_t^{(i)}\right)_{-} + \beta_{t}^{(i)} 
%\end{align}
where $(\cdot)_{+}=\max(\cdot,0)$ and $(\cdot)_{-}=\max(-\cdot,0)$ denote the positive and negative parts respectively. The inventory dynamics is summarized in Figure \ref{fig:inventory_dynamics} above.

\begin{remark} \label{rem:clusters} (Product clusters) In practice, the different items of a warehouse are naturally classified into several groups according to their storage conditions: size, weight and specific climatic conditions, e.g., temperature, pressure, humidity levels, ventilation and light. 
\end{remark}

\begin{remark} \label{rem:capacity} (Storage capacity) In the case where the total storage space is shared among all items ($K=1$), the capacity constraint is fulfilled as soon as $x_t^T \un_n \leq \Lambda$ where $\Lambda$ is the maximum capacity of the warehouse. To model different volumes of items, one can simply consider a vector of unit volumes $v \in \rset^n$ where $v^{(i)}$ is the unit volume of product $i$ and ensures that $x_t^T v \leq \Lambda$ for all $t \geq 0$. Similarly, when dealing with $K$ types of products, the capacity constraint of each subspace $\mathcal{N}_k$ becomes $\sum_{i \in \mathcal{N}_k} x_{t}^{(i)} v^{(i)} \leq \Lambda^{(k)}$ for all $t \geq 0$.
\end{remark}

\begin{remark}\label{rem:overflow}(Storage overflow) Different options are available to address the storage overflow of a product subspace $\mathcal{N}_k$. If all items in $\mathcal{N}_k$ have equal priority levels, one could think of a uniform split of the available storage space of the form $ \Delta_t^{(k)}/|\mathcal{N}_k|$. However, in practice, some items are more important than others and may even be critical for production. This is why, when the storage space overflows, the replenishment quantities should be weighted to favor the replenishment of products non only with low inventory levels but also the ones associated to large shortage costs (see Eq.\eqref{eq:weight})
\end{remark}

\section{MARL for Inventory Management}
\label{sec:marlim}
Recall that the goal of any MARL algorithm is to find a policy $\pi_{\theta}$ maximizing the expected discounted reward $
    V(\pi_{\theta}) = \expec\left[\sum_{t \geq 0} \gamma^t r(s_{t},a_t) | a_t \sim \pi_{\theta} \right].$
The goal of this Section is to derive the necessary formalism, namely the Markov games and associated rewards, in order to apply such MARL methods.

\textbf{Markov games.} The inventory management problem can be modeled as $K$ different Markov games. Consider a fixed product subspace $\mathcal{N}_k$ with $1 \leq k \leq K$. At time step $t \geq 0$, the state of agent $i \in \mathcal{N}_k$ is comprised of: the inventory level $x_t^{(i)}$, the replenishment quantity $\rho_t^{(i)}$, the lead time $\tau_t^{(i)}$ and backlogs $\beta_t^{(i)}$, i.e., $s_t^{(i)}=(x_t^{(i)},\rho_t^{(i)},\tau_t^{(i)},\beta_t^{(i)}) \in \mathcal{S}_k^{i}$. The action of each agent concerns the ordering quantity $a_t^{(i)} \in \mathcal{A}_k^i = \llbracket 0,\Lambda^{(k)} \rrbracket$. The joint state space and action space are respectively given by $\mathcal{S}_k = \prod_{i \in \mathcal{N}_k} S_k^{i}$ and $\mathcal{A}_k = \prod_{i \in \mathcal{N}_k} \mathcal{A}_k^{i}$.  The transition kernel is implicitly defined by all the inventory dynamics equations of Section \ref{subsec:dynamics}.

%For the central controller, $\mathcal{S} = \mathcal{S}_1 \times \ldots \times \mathcal{S}_n$ and $\mathcal{A} = \mathcal{A}_1 \times \ldots \times \mathcal{A}_n$.

\textbf{Costs and rewards.} According to Section \ref{subsec:costs}, at each time step $t \geq 0$ and for a given product subspace $\mathcal{N}_k$, each agent $i \in \mathcal{N}_k$ incurs some inventory costs $C_t^{(i)}$ given by the combination of ordering, holding and shortage costs. In other words, the overall inventory cost may be described as
\begin{align*}
    %\label{eq:cost_single}
    C_t^{(i)} =  \alpha_o \underbrace{a_t^{(i)} C_o^{(i)}}_{\text{ordering}} + \alpha_h \underbrace{x_t^{(i)} C_h^{(i)}}_{\text{holding}} + \alpha_s \underbrace{\beta_{t+1}^{(i)} C_s^{(i)}}_{\text{shortage}}
\end{align*}
where $\alpha_o,\alpha_h,\alpha_s \in [0,1]$ with $\alpha_o+\alpha_h+\alpha_s=1$ are weighting coefficients that translate some expert's knowledge about the desired strategy.
This expression translates the trade-off between the different inventory costs. Hopefully, a trained agent will learn to order low quantities of products that are costly to stock and supply, while maintaining sufficient inventory levels to avoid stock-outs for critical products associated to large shortage costs. This is validated in the numerical experiments of Section \ref{sec:simus}. 

The inventory costs of each agent are associated to the single reward defined as: $r^i(s_t,a_t) = -C_t^i$. Observe that inside a product subspace $\mathcal{N}_k$, the agents are working in a cooperative setting in order to optimize the average reward corresponding to $ r_k(s_{t},a_t) = \sum_{i \in \mathcal{N}_k} r^i(s_t,a_t)/|\mathcal{N}_k|$. This average reward model allows more heterogeneity among agents and facilitates the development of decentralized MARL algorithms \citep{zhang2018fully,doan2019finite}. More subtly, the reward function is sometimes not sharable with others, as the preference may be private to each agent. Note that this setting finds broad applications in engineering systems as sensor networks \citep{rabbat2004distributed}, smart grid \citep{dall2013distributed}, intelligent transportation systems \citep{adler2002cooperative}, and robotics \citep{corke2005networked}.

%Markov games \citep{littman1994markov} and stochastic games \citep{shapley1953stochastic}, constrained MDP \citep{altman1999constrained}.

%\textbf{Demand and Lead-times distributions.} 
%fast-moving items \citep{silver1985decision,burgin1975gamma}
%The randomness of the problem comes from both the stochastic demands and the lead-times. Denote by $\mu_{\delta}$ and $\mu_{\tau}$ the probability distributions of the demand and the lead-times respectively. In practice, these distributions are unknown and may come from general mixtures of discrete probability distributions. These distributions along with the equations of the inventory dynamics define a markovian transition kernel.
\section{Numerical Experiments}
\label{sec:simus}
This Section is dedicated to numerical experiments on real data. Various scenarios regarding the capacity constraints of the warehouses are considered: \textit{(i)} when the items in a product cluster have their own capacity constraints then it is enough to train a single RL agent and apply the resulting behavior to all the items in that cluster; \textit{(ii)} when the items compete for storage space then one may apply some MARL algorithm to deal with the interdependency of items.

\textbf{Stochastic Model.} The demands and lead-times are assumed to be stochastic with stationary distributions. The lead-times follow geometric distributions and the demands follow a mixed law of Poisson process with zero plateaus. More precisely, for each item $i$, the lead-time is given by $\tau^{(i)} \sim \mathcal{G}(p_i)$ with parameter $p_i \in (0,1)$ and the demand distribution is given by
\begin{align*}
 \delta^{(i)} = \left\{
    \begin{array}{ll}
        \mathcal{P}(\mu_i) & \text{with probability } b_i \\
        0 &  \text{with probability } 1-b_i
    \end{array}
\right.
\end{align*}
In other words, the demand process of item $i$ is $\delta^{(i)} \sim X_iY_i$ with $X_i \sim \mathcal{B}(b_i)$ and $Y_i \sim \mathcal{P}(\mu_i)$. 
%The means and variance of the demands are given by
%\begin{align*}
%    \expec[\delta^{(i)}] &= \expec[X_i Y_i] = b_i \mu_i \\
%    \var[\delta^{(i)}] &= \var[X_i Y_i] = b_i \mu_i + b_i(1-b_i)\mu_i^2
%\end{align*}

\textbf{Real data.} The data comes from a company with warehouses in different countries. The data is used to find the model parameters of the stochastic distributions using some maximum likelihood estimators \citep{fisher1922mathematical,white1982maximum,owen2001empirical}. For a fixed item $i \in \mathcal{N}$ with historical data of demands $(\delta_t^{(i)})_{t=1,\ldots,N_i}$ and lead-times $(\tau_t^{(i)})_{t=1,\ldots,T_i}$, the MLE estimators $\hat b_i, \hat \mu_i$ and $\hat p_i$ are given by the empirical averages
\begin{align*}
    \hat b_i = \frac{1}{N_i} \sum_{t=1}^{N_i} \un_{\{\delta_t^{(i)}>0\}}, \qquad  
    \hat \mu_i = \frac{\sum_{t=1}^{N_i} \delta_t^{(i)} \un_{\{\delta_t^{(i)}>0\}}}{\sum_{t=1}^{N_i} \un_{\{\delta_t^{(i)}>0\}}}, \qquad
     \hat p_i = \frac{T_i}{\sum_{t=1}^{T_i} \tau_t^{(i)}}.
\end{align*}
%Consider a fixed item $i \in \mathcal{N}$ with historical data of demands $(\delta_t^{(i)})_{t=1,\ldots,N_i}$. The model parameters $b_i$ and $\mu_i$ of the demands are estimated using the MLE estimators $\hat b_i$ and $\hat \mu_i$ defined as the empirical averages
%\begin{align*}
%    \hat b_i = \frac{1}{N_i} \sum_{t=1}^{N_i} \un_{\{\delta_t^{(i)}>0\}}, \quad  
%    \hat \mu_i = \frac{\sum_{t=1}^{N_i} \delta_t^{(i)} \un_{\{\delta_t^{(i)}>0\}}}{\sum_{t=1}^{N_i} \un_{\{\delta_t^{(i)}>0\}}}.
%\end{align*}
%Similarly, using a data history of lead-times $(\tau_t^{(i)})_{t=1,\ldots,T_i}$, the model parameter $p_i$ is estimated using $\hat p_i$ defined by
%\begin{align*}
% \hat p_i = \frac{T_i}{\sum_{t=1}^{T_i} \tau_t^{(i)}}.
%\end{align*}
For the case study, the total number of products is equal to $n=50$ with time steps equal to months. The history is ranging from $2002$ to $2021$ on different warehouses leading to a maximum number of historical data points equal to $N_{\max}=420$. The horizon is set to $T=240$ months and the methods are tested over $100$ replications (independent random seeds). The different parameter values of each item are summarized in Tables \ref{tab:products_0}-\ref{tab:products_4} in Appendix \ref{app:item_parameters}. %For ease of reproducibility, the code is available upon request.

\begin{figure*}
    \centering
\includegraphics[width=\textwidth/3]{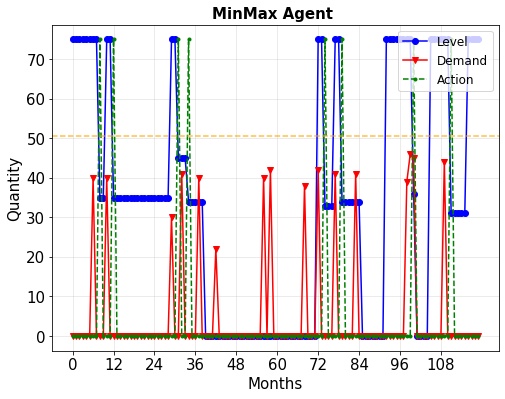}\hfill
\includegraphics[width=\textwidth/3]{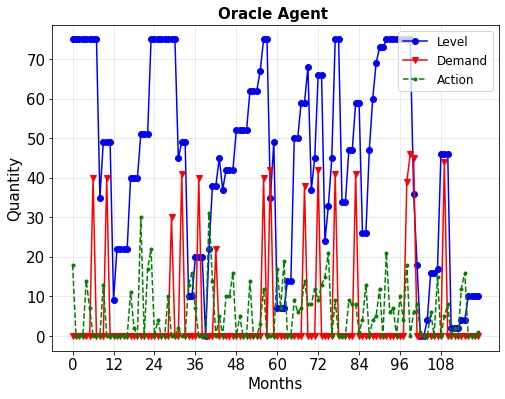}\hfill
\includegraphics[width=\textwidth/3]{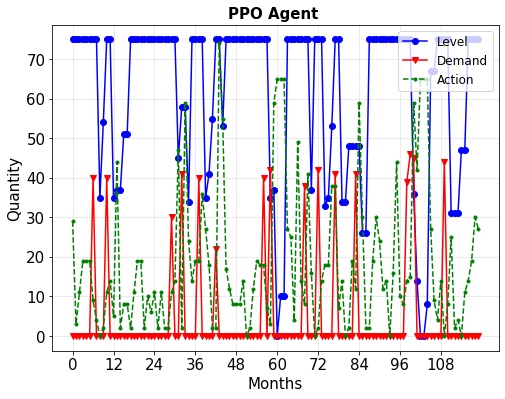}
%\vspace{-0.8cm}
    \caption{Inventory level (\textit{blue}) of an item (ID=$21$) over $T=120$ months with different agents: MinMax (left), Oracle (center), PPO (right). The demand is plotted in \textit{red} and the order actions are plotted in \textit{green}. The safety stock MinMax agent is displayed in \textit{orange}.}
    \label{fig:comparison_actions}
\end{figure*}

\textbf{Agents battle.} In order to evaluate the performance of the developed approach, different baselines are implemented along with the reinforcement learning methods.

\textbullet \ \textit{MinMax agents}: a standard min-max strategy $(s,S)$ from operation research. For these agents, each item has a safety stock $\kappa = \Phi^{-1}(\alpha) \sqrt{(\mu_\tau \sigma_{\delta}^2)  + (\mu_{\delta} \sigma_\tau)^2}$, where $\Phi$ is the \textit{c.d.f.} of $\mathcal{N}(0,1)$, $\alpha \in \{0.90,0.95,0.99\}$ represents the target service level and $\mu,\sigma^2$ are the mean and variance of the demands and lead-times. The value of the service level is set to the classical value $\alpha=0.90$ and the means and variances are computed using the MLE estimators. % and the closed form formulas $\expec[\delta^{(i)}]= b_i \mu_i,  \var[\delta^{(i)}] = b_i \mu_i + b_i(1-b_i)\mu_i^2.$
%The means and variance of the demands are given by
%\begin{align*}
%    \expec[\delta^{(i)}] &= \expec[X_i Y_i] = b_i \mu_i, \\
%    \var[\delta^{(i)}] &= \var[X_i Y_i] = b_i \mu_i + b_i(1-b_i)\mu_i^2.
%\end{align*}
As soon as the inventory level goes below the safety stock, the controller orders the maximum capacity of the corresponding item. Note that such approach is easy to implement but may fail to anticipate spikes in the demand signal.

\textbullet \ \textit{Oracle agents}: such agents implement the following heuristic: given the knowledge of the mean $\mu_{\delta}$ and variance $\sigma_{\delta}^2$ of the demand signal, it is natural to order, at each timestep, according to a random law with mean $\mu_{\delta}$ and variance $\sigma_{\delta}^2$. More precisely, the oracle agents have access to the estimated mean $\hat \mu_{\delta}$ and standard deviation $\hat \sigma_{\delta}$ of their associated products and order at each time according to a normal law $\mathcal{N}(\hat \mu_{\delta},\hat \sigma_{\delta}^2)$ which is clamped to fit the bounds of the action space.  

\textbullet \ \textit{MARL agents}: the reinforcement learning agents are trained using Proximal Policy Optimization (PPO) algorithms \citep{schulman2017proximal} which are stable and effective policy gradient methods. When working with capacity constraints per item, both discrete (Eq. \eqref{eq:gibbs}) and continuous (Eq.\eqref{eq:gaussian}) policies are considered, denoted by PPO-D and PPO-C respectively. When the items compete for storage space, IPPO \citep{tan1993multi} which is an independent version of PPO for multi-agent frameworks is implemented. The different parameter configurations are given in Appendix \ref{app:ppo_details}.

%$[\mu_{\delta}-\sigma_{\delta}; \mu_{\delta}+\sigma_{\delta}]$.

%$C_h = 0.05-0.15 C_o, C_s = 10-15 C_o$ uniformly random

\textbf{Results.} In the idea that similar items among a cluster share some intrinsic features, the question of generalization for a single agent is raised. For that matter, first consider a scenario with a cluster of $5$ items with their own capacity constraints and a single RL agent trained on an average item, \textit{i.e.}, a "virtual" item whose features are given by taking the mean of the features of all the items in that cluster is trained. Then the learned behavior on the $5$ items of the cluster is tested and the different cumulative costs are reported. Table \ref{tab:results_single_agents} below presents the average cumulative costs in \$ obtained over $100$ replications of the different methods for the $5$ items over an horizon of $T=240$ months. 

\vspace{-0.2cm}
\begin{table}[h]
    \centering
    %\parbox{.48\linewidth}{
    \begin{tabular}{|c|r|r|r|r|}
       \hline
       ID & \multicolumn{1}{c|}{MinMax} & \multicolumn{1}{c|}{Oracle} & \multicolumn{1}{c|}{PPO-D} & \multicolumn{1}{c|}{PPO-C} \\
    \hline
    0 & 48,554,986 & 10,183,088 & 4,863,202 & \textbf{4,616,016} \\ % 10.5
    \hline
    1 & 52,993,931 & 16,917,389 & 6,865,991 & \textbf{6,385,378} \\ %8.3
    \hline
    2 & 70,467,282 & 21,426,806 & 8,727,215 & \textbf{8,087,258} \\
    \hline
    3 & 72,220,832 & 12,722,887 & 9,854,047 & \textbf{5,280,345} \\
    \hline
    4 & 79,235,272 & 16,976,630 & 8,801,628 & \textbf{4,808,191} \\
    \hline
    \end{tabular}
    \caption{Average Cumulative cost in \$ over 100 replications for different items over $T=240$ months.}
    \label{tab:results_single_agents}
    \end{table}
    \vspace{-1.2cm}
    \begin{table}
    %\parbox{.35\linewidth}{
    \centering
    \begin{tabular}{|c|c|c|c|c|c|}
    \hline
       ID & 0 & 1 & 2 & 3 & 4  \\
   \hline
   \hline
    MinMax & 39 & 41 & 42 & 51 & 49 \\
    \hline
    Oracle & 10 & 11 & 13 & 8 & 10 \\
    \hline
    PPO-D & 0 & 0 & 0 & 0 & 0 \\
    \hline 
    PPO-C & 0 & 0 & 0 & 0 & 0 \\
    \hline
    \end{tabular}
    \caption{Average Item Shortage over 100 replications for different items over $T=240$ months.}
    \label{tab:shorts_single_agents}
\end{table}

\iffalse
\begin{table}[h]
    \centering
    \begin{tabular}{|c|r|r|r|r|}
       \hline
       ID & \multicolumn{1}{c|}{MinMax} & \multicolumn{1}{c|}{Oracle} & \multicolumn{1}{c|}{PPO-D} & \multicolumn{1}{c|}{PPO-C} \\
    \hline
    0 & 48,554,986 & 10,183,088 & 4,863,202 & \textbf{4,616,016} \\ % 10.5
    \hline
    1 & 52,993,931 & 16,917,389 & 6,865,991 & \textbf{6,385,378} \\ %8.3
    \hline
    2 & 70,467,282 & 21,426,806 & 8,727,215 & \textbf{8,087,258} \\
    \hline
    3 & 72,220,832 & 12,722,887 & 9,854,047 & \textbf{5,280,345} \\
    \hline
    4 & 79,235,272 & 16,976,630 & 8,801,628 & \textbf{4,808,191} \\
    \hline
    \end{tabular}
    \caption{Average Cumulative cost in \$ over 100 replications for different items over an horizon of $T=240$ months.}
    \label{tab:results_single_agents}
\end{table}
\fi
\newpage
%\vspace{-3.cm}
Regarding the average cumulative costs, the clear winners are the RL-based methods. Indeed, the PPO methods are statistically better than the two other baselines with a cost reduction factor ranging from $8$ (for item ID=1) to $16$ (for item ID=4) compared to the standard $(s,S)$-strategy. Interestingly, among the PPO methods, the one with continuous actions presents the best performance. Another important result concerns the number of items shortages as shown by Table \ref{tab:shorts_single_agents} above. % presents the average item shortages obtained over $100$ replications of the different methods for the $5$ items over an horizon of $T=240$ months. 
%Observe that, also for this criterion, 
Once again, the RL-based methods outperform the two standard baselines. Furthermore, the PPO agents learned an optimal strategy in terms of stock-outs since the number of item shortages is always equal to zero. Such careful behavior is confirmed by Figure \ref{fig:comparison_actions} where the evolution of the inventory levels is plotted, over an horizon of $T=120$ months, for a particular item according to the different strategies. Note that compared to the MinMax and Oracle agents, the inventory levels of the PPO agent are always above the demand signal which ensure the avoidance of stock-outs. Interestingly, the RL agents have the nice following interpretation: whenever there is either a demand spike or a demand plateau for a long time, there is an incentive to order. Additional numerical results concerning all $50$ different items are available in Appendices \ref{app:costs_single} and \ref{app:shortages_single} for the average cumulative costs and the average item shortages respectively.
\begin{figure*}
    \centering
\includegraphics[width=\textwidth/3]{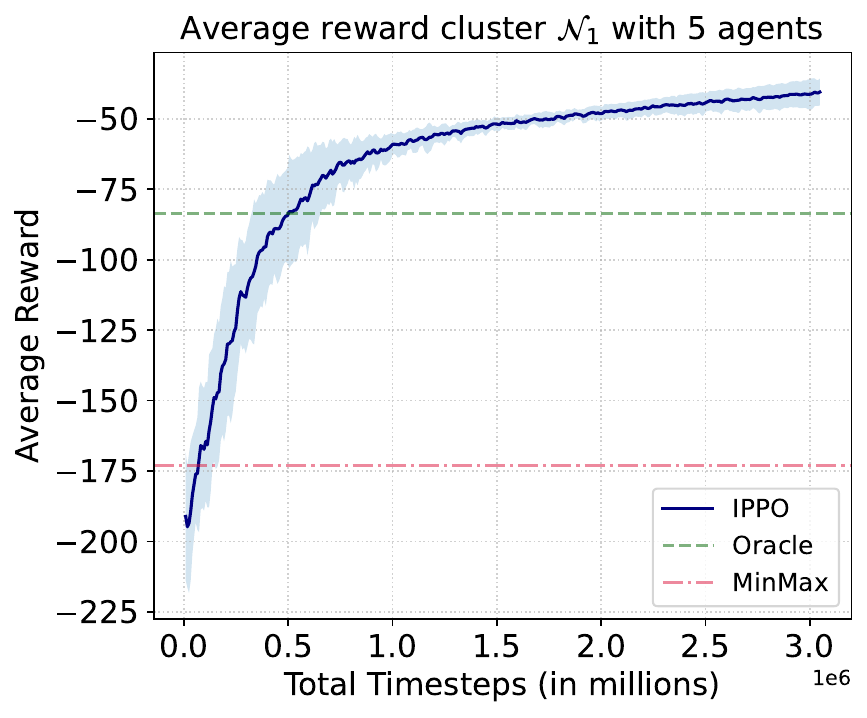}\hfill
\includegraphics[width=\textwidth/3]{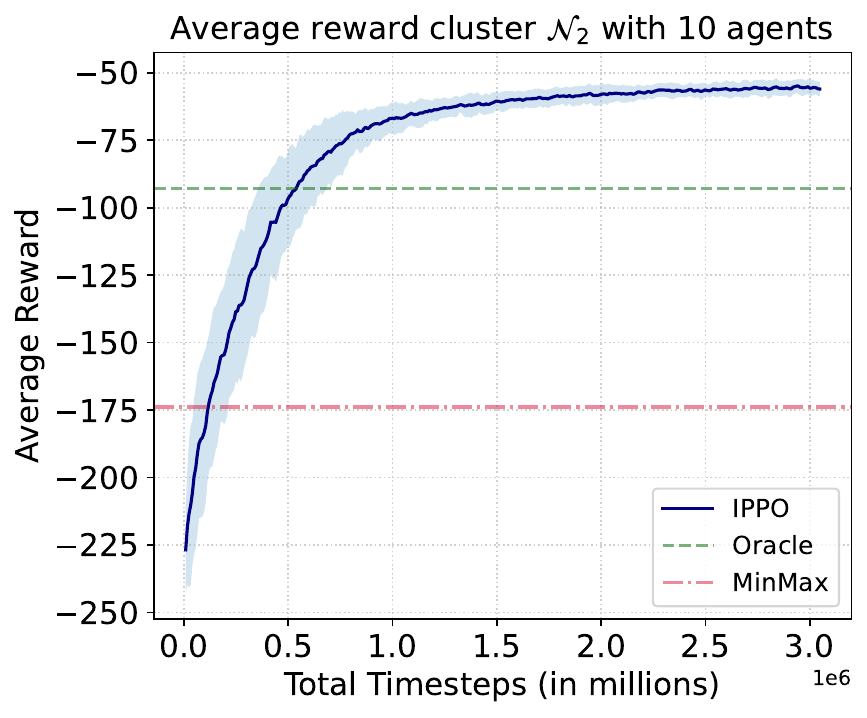}\hfill
\includegraphics[width=\textwidth/3]{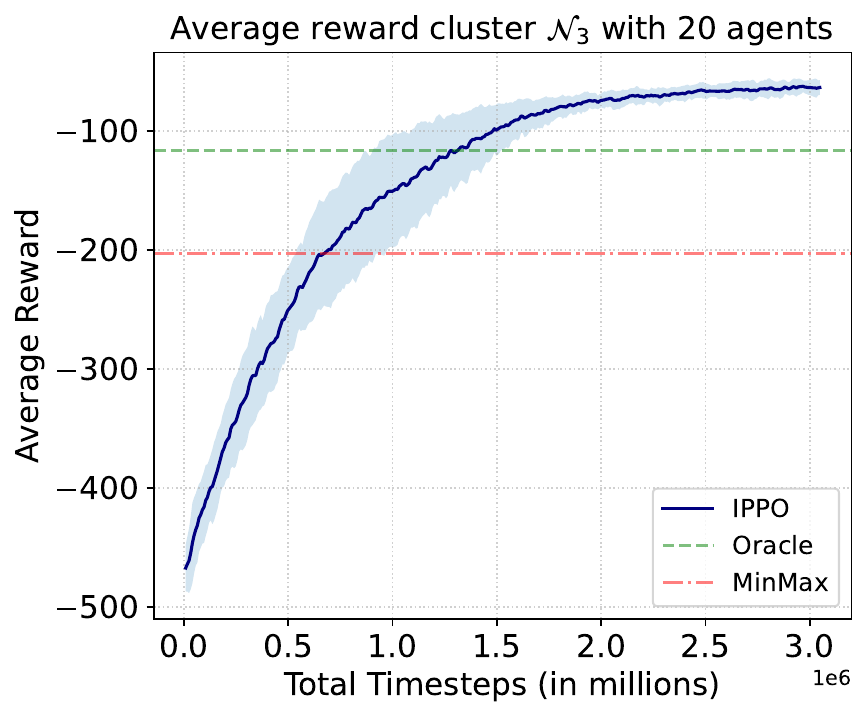}
\vspace{-0.5cm}
    \caption{Learning curves for the different clusters $\mathcal{N}_1,\mathcal{N}_2$ and $\mathcal{N}_3$ where the mean and standard deviation of IPPO are plotted in \textit{blue} and the horizontal lines represent the average reward for the two baselines Oracle (\textit{green}) and MinMax (\textit{red}) computed over $100$ replications.
    \vspace{-0.5cm}}
    \label{fig:learning_curves}
\end{figure*}

Another scenario involving storage capacity constraint is considered with three different clusters $\mathcal{N}_1, \mathcal{N}_2$ and $\mathcal{N}_3$ composed of $5, 10$ and $20$ items respectively. The last cluster represents a more complex task as it involves many agents. As before, the different methods based on the average cumulative cost over a cluster and the average item shortages over a cluster are compared, both over an horizon of $T=240$ months. For each cluster, one may take a look at the different training curves of the MARL agents. Such training is repeated on $10$ replications and the results are displayed in Figure \ref{fig:learning_curves} which gathers the evolution of the normalized average reward along the training episodes. The means and standard deviations of the average reward of the MARL agents are plotted in blue. For ease of comparison, two horizontal lines corresponding to the normalized average reward of the MinMax and Oracle agents are added. Thus, it allows to check the performance gain obtained with the RL-based methods compared to standard baselines. Observe that for clusters $\mathcal{N}_1$ and $\mathcal{N}_2$, the MARL agents quickly outperform the baselines in approximately $500,000$ timesteps but need more than 1M timesteps on cluster $\mathcal{N}_3$ to surpass the Oracle agents. Concerning the test performance of the learned behaviors, the different Tables below show the average cumulative cost and average stock-outs over the different items, obtained over $100$ replications and an horizon of $T=240$ months.

\begin{table}[h]
    \centering
    \parbox{.5\linewidth}{
    \begin{tabular}{|c|c|c|c|}
       \hline
       Cluster & $\mathcal{N}_1$ & $\mathcal{N}_2$ & $\mathcal{N}_3$ \\
    \hline
    MinMax & 49,219,195 & 55,148,027 & 63,319,299 \\
    \hline
    Oracle & 29,596,024 & 31,047,144 & 38,002,027 \\
    %\hline
    %PPO-D & 25,206,893 & ??? & ??? \\
    \hline
    IPPO-C & \textbf{12,048,666} & \textbf{7,905,165} & \textbf{14,239,567} \\
    \hline
    \end{tabular}
    \caption{Average Cumulative cost in \$ over different clusters, computed over 100 replications for an horizon of $T=240$ months.}
    \label{tab:results_MARL}}
    \hfill
    \parbox{.44\linewidth}{
    \centering
    \begin{tabular}{|c|c|c|c|}
       \hline
      Cluster & $\mathcal{N}_1$ & $\mathcal{N}_2$ & $\mathcal{N}_3$ \\
    \hline
    MinMax & 18 & 18 & 23 \\
    \hline
    Oracle & 13 & 9 & 11\\
    %\hline
    %PPO-D & 5 & & ???\\
    \hline
    IPPO-C & \textbf{3} & \textbf{0} & \textbf{2} \\
    \hline
    \end{tabular}
    \caption{Average Item shortages computed over 100 replications for an horizon of $T=240$ months.}
    \label{tab:shorts_MARL}}
\end{table}

\iffalse
\begin{table}[h]
    \centering
    \begin{tabular}{|c|c|c|c|}
       \hline
       Cluster & $\mathcal{N}_1$ & $\mathcal{N}_2$ & $\mathcal{N}_3$ \\
    \hline
    MinMax & 49,219,195 & 55,148,027 & 63,319,299 \\
    \hline
    Oracle & 29,596,024 & 31,047,144 & 38,002,027 \\
    %\hline
    %PPO-D & 25,206,893 & ??? & ??? \\
    \hline
    IPPO-C & \textbf{12,048,666} & \textbf{7,905,165} & \textbf{14,239,567} \\
    \hline
    \end{tabular}
    \caption{Average Cumulative cost in \$ over different clusters, computed over 100 replications for an horizon of $T=240$ months.}
    \label{tab:results_MARL}
\end{table}
\fi
%\vspace{-0.2cm}
Once again, the MARL-based agents present the best performance, both in terms of cost savings (see Table \ref{tab:results_MARL}) and avoidance of shortages (see Table \ref{tab:shorts_MARL}). Observe that for cluster $\mathcal{N}_1$, the MARL agents allows an overall cost reduction of $75\%$ compared to the standard $(s,S)$ strategy, $85\%$ for cluster $\mathcal{N}_2$ and about $78\%$ for cluster $\mathcal{N}_3$. For each cluster, the details about the average costs and shortages of all items are available in Appendix \ref{app:costs_multi}.

\iffalse
\begin{table}[h]
    \centering
    \begin{tabular}{|c|c|c|c|}
       \hline
      Cluster & $\mathcal{N}_1$ & $\mathcal{N}_2$ & $\mathcal{N}_3$ \\
    \hline
    MinMax & 18 & 18 & 23 \\
    \hline
    Oracle & 13 & 9 & 11\\
    %\hline
    %PPO-D & 5 & & ???\\
    \hline
    IPPO-C & \textbf{3} & \textbf{0} & \textbf{2} \\
    \hline
    \end{tabular}
    \caption{Average Item shortages over different product clusters, computed over 100 replications for an horizon of $T=240$ months.}
    \label{tab:shorts_MARL}
\end{table}
\fi

\section{Conclusion \& Discussion}
\label{sec:conclusion}

Maintaining the right balance between the supply and demand of products by optimizing replenishment decisions is one of the most important challenges for inventory management systems. In this paper, a rigorous methodological and practical reinforcement learning framework to address the inventory management problem for a single-echelon multi-products supply chain on a production line with stochastic demands and lead-times has been developed. The method has been illustrated with extensive numerical experiments on real data for both single and multi-agents algorithms.

This problem is very close to real-world use cases. Not only does it handle stochastic demands and lead-times but is designed for deployment as an actual business solution on production lines. Future work will focus on the statistical properties of the developed framework by further exploring links with mean-field approximation theory and the effects of exogenous variables in reinforcement learning. 

\section*{Acknowledgements}
The authors report financial support was provided by TotalEnergies SE. The authors have patent issued to TotalEnergies SE.

\bibliographystyle{apalike}
\bibliography{main}

%\clearpage
\onecolumn

\begin{center}
\LARGE
    \textbf{Appendix}
\end{center}

Appendix \ref{app:ppo_details} collects the technical details related to the implementation of Proximal Policy Optimization algorithms, namely the different loss functions and the associated hyper-parameters used for the training phase. Appendix \ref{app:item_parameters} gathers the different item parameters of the real data. Appendices \ref{app:costs_single} and \ref{app:shortages_single} present additional results concerning the cumulative costs and item shortages for single agents. Similarly, Appendix \ref{app:costs_multi} is dedicated to the detailed numerical results of MARL based methods.

\appendix
\section{Proximal Policy Optimization algorithms}
\label{app:ppo_details}

\subsection{PPO methodology and model}

PPO is a model-free on-policy RL algorithm that works well for both discrete and continuous action space environments. PPO utilizes an actor-critic framework, where there are two networks, an actor (policy network) and critic network (value function). Such algorithms are part of the family of policy gradient algorithms which use a parameterized action-selection policy $\pi_{\theta}$ with $\theta \in \Theta \subset \rset^d$ and update the policy parameter $\theta$ on each step in the direction of an estimate of the gradient of the performance with respect to the policy parameter: $$\theta \leftarrow \theta + \eta \hat g,$$  
where $\eta$ is the learning rate, $\hat g = \expec_t[\nabla_{\theta} \log \pi_{\theta}(a_t|s_t) \hat A_t]$ is a gradient estimate, $\hat A_t$ is an estimator of the advantage function at timestep $t$ and the expectation $\expec_t$ indicates the empirical average over a finite batch of samples, in an algorithm that alternates between sampling and optimization.

The implementation of the Proximal Policy Optimization algorithms follows the one of the open-source library RLlib \citep{liang2018rllib}. Whereas standard policy gradient methods perform one gradient update per data sample, PPO enables multiple epochs of minibatch updates. According to \cite{schulman2017proximal}, compared to TRPO methods, PPO algorithms are much simpler to implement, more general, and have better sample complexity (empirically). PPO’s clipped objective supports multiple SGD passes over the same batch of experiences and RLlib’s PPO can scale out using multiple workers for experience collection, and also with multiple GPUs for SGD.

Denote by $r_t(\theta) = \pi_{\theta}(a_t|s_t)/\pi_{\theta_{\text{old}}}(a_t|s_t)$ the probability ratio between old and new policies so that TRPO methods aim at optimizing the following surrogate objective: $\mathcal{L}_{\text{TRPO}}(\theta) = \expec_t[r_t(\theta) \hat A_t]$. In comparison, PPO methods rely on the following clipped objective function: $$\mathcal{L}_{\text{actor}}(\theta) = \expec_t\left[\min\left\{r_t(\theta)\hat A_{t}, \text{clip}(r_t(\theta),1-\varepsilon,1+\varepsilon)\hat A_{t}\right\} \right].$$ 

The motivation for this objective is as follows. The second term inside the min modifies the surrogate objective by clipping the probability ratio, which removes the incentive for moving $r_t$ outside of the interval $[1 - \varepsilon, 1 + \varepsilon]$. By taking the minimum of the clipped and unclipped objective, the final objective is a lower bound (i.e., a pessimistic bound) on the unclipped objective. With this
scheme, PPO methods only ignore the change in probability ratio when it would make the objective improve, and include it when it makes the objective worse. 
This very idea of clipping can also be applied to the critic whose aim is to approximate the value function using some neural network, leading to the following critic loss
$$\mathcal{L}_{\text{critic}}(\theta) = \expec_t\left[\max\left\{
(V_{\theta}(s_t)-\hat V_t)^2,
(V_{\theta_{\text{old}}}(s_t) -\hat V_t + \text{clip}(V_{\theta}(s_t)-V_{\theta_{\text{old}}}(s_t),-\varepsilon,+\varepsilon))^2
\right\} \right]$$
Furthermore, in addition to clipped surrogate objective, consider some KL divergence and entropy terms. For the KL term, this follows from the fact that a certain surrogate objective (which computes the max KL over states instead of the mean) forms a lower bound (i.e., a pessimistic bound) on the performance of the current policy. Finally, the total objective can be augmented by adding
an entropy bonus $\mathcal{H}$ to ensure sufficient exploration, as suggested in past work \citep{mnih2015human}. Overall, the goal is to maximize the following objective function
\begin{align*}
\mathcal{L}(\theta) = \mathcal{L}_{\text{actor}}(\theta) - c_1 \mathcal{L}_{\text{critic}}(\theta) - c_2 \expec_t[ D_{KL}(\pi_{\theta_{\text{old}}}(\cdot|s_t),\pi_{\theta}(\cdot|s_t))] +  c_3 \expec_t[\mathcal{H}(\pi_\theta)(s_t)],
\end{align*}
where $\mathcal{L}_{\text{actor}}$ is the policy loss, $\mathcal{L}_{\text{critic}}$ is the critic loss, $D_{KL}$ is the Kullback-Liebler divergence and $\mathcal{H}$ is the entropy bonus. The constants $c_1,c_2$ and $c_3$ are the value-function loss coefficient, the KL coefficient and the entropy coefficient. 

\newpage
\subsection{Implementation Details}

When working with storage constraints per item, it is enough to only train a few agents. The $50$ items of the different warehouses are split in $2$ groups according to the median of lead-time in order to train only $2$ average RL agent. Such agents are then tested on all the items in the corresponding group. In order to compare the effect of working with discrete or continuous actions spaces, two RL agents are trained using Discrete actions (PPO-D1 and PPO-D2), and 2 RL agents using Continuous actions (PPO-C1 and PPO-C2). When working with items competing for storage space, MARL algorithms are used, suchg as IPPO and other variations with shared critic. Tables \ref{tab:common_params} and \ref{tab:specific_params} gather all the details about the different hyper parameters used for the traning of (MA)RL agents. In all the experiments, the following parameters are fixed:

\begin{table}[h!]
    \centering
    \begin{tabular}{lc}
    \hline
    Parameter & Value \\
    \hline
    HORIZON & 200 \\
    \hline
    GAMMA & 0.99 \\ 
    \hline
    LEARNING RATE & 1e-4 \\
    \hline
    VF SHARE LAYERS & False \\
    \hline
    ROLLOUT FRAGMENTLENGTH & 200 \\
    \hline
    BATCHMODE & complete \\
    \hline
    TRAIN BATCH SIZE & 8000 \\
    \hline
    SGD MINIBACTH SIZE & 250 \\
    \hline
    NUM SGD ITER & 20 \\
    \hline
    NORMALIZE ACTIONS & True \\
    \hline
    FCNET ACTIVATION & relu \\
    \hline
    USE CRITIC & True  \\
    \hline
    GAE LAMBDA & 1 \\
    \hline
    KL COEFF & 2e-1 \\
    \hline
    KL TARGET & 1e-2 \\
    \hline
    ENTROPY COEFF & 0.01 \\
    \hline
    CLIP PARAM & 0.3 \\
    \hline
    \end{tabular}
    \caption{Hyper parameters Details for Training}
    \label{tab:common_params}
\end{table}

The following parameters are specific to the different RL agents:
\begin{table}[h!]
    \centering
    \begin{tabular}{lcccc}
    \hline
      Parameter &  PPO-D1 & PPO-D2 & PPO-C1 & PPO-D2 \\
    \hline
    FCNET HIDDEN & [512, 512] & [512, 512] & [512, 512]  & [512, 512]   \\
    \hline
    GRAD CLIP & 40 & 40 & 40 & 40  \\
    \hline
    LEARNING RATE & 1e-4 & 1e-4 & 1e-4 & 2e-4  \\
    \hline
    VF SHARE LAYERS & False & False & False & False \\
    \hline
    USE GAE & True & True &  True & True \\
    \hline
    VF CLIP PARAM & 1e3 & 1e4 & 1e3 & 5e2  \\
    \hline
    VF LOSS COEFF & 1 & 1e-2 & 1e-2 & 1e-2  \\
    \hline
    \end{tabular}
    \caption{Hyper parameters Details for Training}
    \label{tab:specific_params}
\end{table}

\begin{table}[h!]
    \centering
    \begin{tabular}{lccc}
    \hline
      Parameter &  IPPO-$\mathcal{N}_1$ & IPPO-$\mathcal{N}_2$ &  IPPO-$\mathcal{N}_3$ \\
    \hline
    FCNET HIDDEN &  [512, 256] & [512,256] & [512,256] \\
    \hline
    GRAD CLIP & 40 & 20 & 20 \\
    \hline
    LEARNING RATE &  5e-5 & 2e-5 & 2e-5 \\
    \hline
    VF SHARE LAYERS & True & True & True  \\
    \hline
    USE GAE & False & False & False \\
    \hline
    VF CLIP PARAM & 5e2 & 5e2 & 5e2   \\
    \hline
    VF LOSS COEFF & 1e-3 & 1e-4 & 1e-4   \\
    \hline
    \end{tabular}
    \caption{Hyper parameters Details for Training}
    \label{tab:specific_params_ippo}
\end{table}

\newpage
\section{Item Hyperparameters}
\label{app:item_parameters}

The data is used to find the model parameters of the stochastic distributions using some MLE estimators. For a fixed item $i \in \mathcal{N}$ with historical data of demands $(\delta_t^{(i)})_{t=1,\ldots,N_i}$ and lead-times $(\tau_t^{(i)})_{t=1,\ldots,T_i}$, the MLE estimators $\hat b_i, \hat \mu_i$ and $\hat p_i$ are given by the empirical averages
\begin{align*}
    \hat b_i = \frac{1}{N_i} \sum_{t=1}^{N_i} \un_{\{\delta_t^{(i)}>0\}} \qquad  
    \hat \mu_i = \frac{\sum_{t=1}^{N_i} \delta_t^{(i)} \un_{\{\delta_t^{(i)}>0\}}}{\sum_{t=1}^{N_i} \un_{\{\delta_t^{(i)}>0\}}} \qquad
     \hat p_i = \frac{T_i}{\sum_{t=1}^{T_i} \tau_t^{(i)}}
\end{align*}
The different Tables below summarize the parameter values for each item.

\begin{table}[h!]
    \centering
    \begin{tabular}{ccccccc}
    \hline
        ID & $\hat b$ & $\hat \mu$ & $\hat p$ & $C_o$ & $C_h$ & $C_s$ \\
    \hline
      0 & 0.33 & 6.23 & 0.12 & 1,010 & 57 & 11,097 \\
    \hline
    1 & 0.12 & 17.33 & 0.17 & 1,092 & 125 & 11,800 \\
    \hline
    2 & 0.21 & 11.0 & 0.17 & 1,363 & 159 & 14,887 \\
    \hline
    3 & 0.24 & 9.04 & 0.11 & 1,125 & 131 & 12,881 \\
    \hline
    4 & 0.17 & 12.0 & 0.11 & 1,007 & 119 & 14,758 \\
    \hline
    5 & 0.31 & 6.87 & 0.11 & 1,174 & 65 & 15,954 \\
    \hline
    6 & 0.17 & 12.5 & 0.12 & 1,280 & 104 & 18,109 \\
    \hline
    7 & 0.12 & 17.25 & 0.12 & 1,220 & 71 & 14,450 \\
    \hline
    8 & 0.18 & 11.82 & 0.19 & 2,250 & 269 & 23,984 \\
    \hline
    9 & 0.29 & 8.46 & 0.13 & 1,356 & 129 & 13,998 \\
    \hline
    \end{tabular}
    \caption{Products Parameters and Costs for Items 0-9}
    \label{tab:products_0}
\end{table}

\begin{table}[h!]
    \centering
    %\parbox{.45\linewidth}{
    \begin{tabular}{ccccccc}
    \hline
        ID & $\hat b$ & $\hat \mu$ & $\hat p$ & $C_o$ & $C_h$ & $C_s$ \\
    % \hline
    \hline
    10 & 0.29 & 8.08 & 0.15 & 1,597 & 170 & 21,512 \\
    \hline
    11 & 0.11 & 22.0 & 0.14 & 1,069 & 100 & 14,184 \\
    \hline
    12 & 0.4 & 8.25 & 0.2 & 1,020 & 112 & 15,244 \\
    \hline
    13 & 0.17 & 16.25 & 0.15 & 1,342 & 149 & 14,059 \\
    \hline
    14 & 0.15 & 20.59 & 0.12 & 1,080 & 112 & 11,352 \\
    \hline
    15 & 0.08 & 24.0 & 0.41 & 3,380 & 298 & 37,941 \\
    \hline
    16 & 0.24 & 9.83 & 0.1 & 1,857 & 194 & 23,874 \\
    \hline
    17 & 0.12 & 26.67 & 0.14 & 1,042 & 107 & 11,000 \\
    \hline
    18 & 0.12 & 20.0 & 0.1 & 1,360 & 174 & 17,718 \\
    \hline
    19 & 0.4 & 7.07 & 0.12 & 1,690 & 215 & 20,403 \\
    \hline
    \end{tabular}
    \label{tab:products_1}
    \caption{Products Parameters and Costs for Items 10-19}
    %}
    \hfill
    %\parbox{.45\linewidth}{
\end{table}
\begin{table}[h!]
    \centering
    \begin{tabular}{ccccccc}
    \hline
        ID & $\hat b$ & $\hat \mu$ & $\hat p$ & $C_o$ & $C_h$ & $C_s$ \\
    % \hline
    \hline
    20 & 0.08 & 40.0 & 0.15 & 1,110 & 95 & 12,873 \\
    \hline
    21 & 0.38 & 9.5 & 0.12 & 1,270 & 181 & 13,578 \\
    \hline
    22 & 0.08 & 32.67 & 0.1 & 1,276 & 184 & 16,441 \\
    \hline
    23 & 0.17 & 23.58 & 0.2 & 1,170 & 60 & 15,026 \\
    \hline
    24 & 0.14 & 20.4 & 0.11 & 2,084 & 270 & 25,725 \\
    \hline
    25 & 0.08 & 40.0 & 0.12 & 1,371 & 177 & 14,804 \\
    \hline
    26 & 0.11 & 32.5 & 0.09 & 1,092 & 152 & 12,305 \\
    \hline
    27 & 0.08 & 32.0 & 0.11 & 2,104 & 135 & 24,015 \\
    \hline
    28 & 0.11 & 36.5 & 0.15 & 1,252 & 139 & 14,186 \\
    \hline
    29 & 0.39 & 7.81 & 0.1 & 1,792 & 93 & 25,279 \\
    \hline
    \end{tabular}
    \caption{Products Parameters and Costs for Items 20-29}
    %}
    \label{tab:products_2}
\end{table}

\begin{table}[h!]
    \centering
    %\parbox{.45\linewidth}{
    \begin{tabular}{ccccccc}
    \hline
        ID & $\hat b$ & $\hat \mu$ & $\hat p$ & $C_o$ & $C_h$ & $C_s$ \\
    \hline
    30 & 0.28 & 15.25 & 0.1 & 1,085 & 77 & 14,038 \\
    \hline
    31 & 0.08 & 40.0 & 0.08 & 1,445 & 164 & 14,742 \\
    \hline
    32 & 0.19 & 18.3 & 0.11 & 2,284 & 304 & 23,957 \\
    \hline
    33 & 0.26 & 19.73 & 0.16 & 1,142 & 138 & 13,926 \\
    \hline
    34 & 0.17 & 26.0 & 0.15 & 1,342 & 196 & 16,559 \\
    \hline
    35 & 0.33 & 10.39 & 0.07 & 1,851 & 267 & 19,626 \\
    \hline
    36 & 0.14 & 26.0 & 0.15 & 1,765 & 225 & 25,052 \\
    \hline
    37 & 0.1 & 35.67 & 0.11 & 2,070 & 130 & 20,910 \\
    \hline
    38 & 0.12 & 30.17 & 0.09 & 1,380 & 204 & 20,550 \\
    \hline
    39 & 0.33 & 6.58 & 0.06 & 4,470 & 456 & 51,326 \\
    \hline
    \end{tabular}
    \caption{Products Parameters and Costs for Items 30-39}
    \label{tab:products_3}
\end{table}

\begin{table}[h!]
    \centering
    %\parbox{.45\linewidth}{
    \begin{tabular}{ccccccc}
    \hline
        ID & $\hat b$ & $\hat \mu$ & $\hat p$ & $C_o$ & $C_h$ & $C_s$ \\
 \hline
    40 & 0.29 & 18.14 & 0.17 & 1,689 & 158 & 22,534 \\
    \hline
    41 & 0.08 & 50.0 & 0.1 & 1,329 & 141 & 17,654 \\
    \hline
    42 & 0.22 & 20.77 & 0.13 & 2,312 & 257 & 29,177 \\
    \hline
    43 & 0.44 & 14.33 & 0.1 & 1,308 & 100 & 15,170 \\
    \hline
    44 & 0.08 & 70.0 & 0.1 & 1,308 & 179 & 13,718 \\
    \hline
    45 & 0.17 & 34.0 & 0.18 & 3,590 & 328 & 38,035 \\
    \hline
    46 & 0.08 & 108.0 & 0.12 & 1,011 & 87 & 10,134 \\
    \hline
    47 & 0.19 & 111.0 & 0.12 & 1,049 & 100 & 15,621 \\
    \hline
    48 & 0.18 & 97.88 & 0.11 & 1,851 & 171 & 20,451 \\
    \hline
    49 & 0.08 & 217.0 & 0.11 & 1,851 & 114 & 25,362 \\
    \hline
    \end{tabular}
    \caption{Products Parameters and Costs for Items 40-49}
    \label{tab:products_4}
    %}
\end{table}

%%%%%%%%%%%%%%%%%%%%%%%%%%%%%%%%%%%%%%%%%%%%%%%%%%%%%%%%%%
\section{Average Cumulative Costs}
\label{app:costs_single}

\subsection{Numerical Results}
\label{subsec:num_results_single}

The different Tables below present the full results of cumulative costs for the numerical experiments with storage space constraints per item. The $50$ items are split in $2$ groups according to the median of lead-time in order to train only $2$ average RL agent. Such agents are then tested on all the items in the corresponding group.
\begin{table}[h]
    \centering
    % \parbox{.45\linewidth}{
    \begin{tabular}{crrrr}
       \hline
       ID & \multicolumn{1}{c}{MinMax} & \multicolumn{1}{c}{Oracle} & \multicolumn{1}{c}{PPO-D} & \multicolumn{1}{c}{PPO-C} \\
    \hline
    0 & 48,554,986 & 10,183,088 & 4,863,202 & \textbf{4,616,016} \\
    \hline
    1 & 52,993,931 & 16,917,389 & 6,865,991 & \textbf{6,385,378} \\
    \hline
    2 & 70,467,282 & 21,426,806 & 8,727,215 & \textbf{8,087,258} \\
    \hline
    3 & 72,220,832 & 12,722,887 & 9,854,047 & \textbf{5,280,345} \\
    \hline
    4 & 79,235,272 & 16,976,630 & 8,801,628 & \textbf{4,808,191} \\
    \hline
    5 & 79,945,152 & 17,793,765 & 9,068,217 & \textbf{4,270,027} \\
    \hline
    6 & 94,856,066 & 23,070,095 & 6,725,252 & \textbf{6,330,215} \\
    \hline
    7 & 94,888,885 & 23,194,601 & 9,540,166 & \textbf{4,599,449} \\
    \hline
    8 & 96,678,620 & 30,082,381 & 15,010,899 & \textbf{13,646,991} \\
    \hline
    9 & 98,917,563 & 15,738,319 & 7,593,699 & \textbf{7,229,608} \\
    \hline
     \end{tabular}
    \caption{Average Cumulative Costs in \$ obtained over $100$ replications, horizon of $T=240$ months for Items 0-9.}
    \label{tab:costs_0_9}
    % }
    % \hfill
    \end{table}
\begin{table}[h]
    \centering
    % \parbox{.45\linewidth}{
     \begin{tabular}{crrrr}
       \hline
       ID & \multicolumn{1}{c}{MinMax} & \multicolumn{1}{c}{Oracle} & \multicolumn{1}{c}{PPO-D} & \multicolumn{1}{c}{PPO-C} \\
    \hline
    10 & 110,582,575 & 24,999,881 & 9,564,232 & \textbf{9,038,083} \\
    \hline
    11 & 112,399,072 & 42,910,634 & 7,574,747 & \textbf{6,628,323} \\
    \hline
    12 & 115,603,447 & 46,823,147 & 6,853,628 & \textbf{6,103,570} \\
    \hline
    13 & 124,832,231 & 39,888,919 & 8,243,508 & \textbf{7,665,907} \\
    \hline
    14 & 152,157,505 & 47,941,527 & 6,445,518 & \textbf{6,418,000} \\
    \hline
    15 & 156,482,922 & 64,905,614 & 26,221,197 & \textbf{18,708,044} \\
    \hline
    16 & 162,805,503 & 32,745,666 & 15,653,937 & \textbf{8,462,765} \\
    \hline
    17 & 165,864,536 & 74,037,166 & 10,165,442 & \textbf{9,631,728} \\
    \hline
    18 & 167,274,515 & 46,875,740 & 12,546,177 & \textbf{7,122,270} \\
    \hline
    19 & 174,895,434 & 56,821,488 & 10,050,155 & \textbf{9,706,447} \\
    \hline
     \end{tabular}
    \caption{Average Cumulative Costs in \$ obtained over $100$ replications, horizon of $T=240$ months for Items 10-19.}
    \label{tab:costs_10_19}
    % }
    \end{table}

\begin{table}[h]
    \centering
    % \parbox{.45\linewidth}{
    \begin{tabular}{crrrr}
       \hline
       ID & \multicolumn{1}{c}{MinMax} & \multicolumn{1}{c}{Oracle} & \multicolumn{1}{c}{PPO-D} & \multicolumn{1}{c}{PPO-C} \\
    \hline
    20 & 182,267,502 & 130,132,350 & \textbf{33,743,915} & 36,077,127 \\
    \hline
    21 & 188,855,483 & 66,602,602 & 8,227,594 & \textbf{7,704,383} \\
    \hline
    22 & 209,851,096 & 101,916,137 & \textbf{18,964,499} & 19,156,603 \\
    \hline
    23 & 240,876,429 & 119,152,696 & 10,058,032 & \textbf{9,181,690} \\
    \hline
    24 & 253,729,788 & 94,653,074 & 18,857,686 & \textbf{14,191,791} \\
    \hline
    25 & 259,717,238 & 154,572,208 & \textbf{39,887,428} & 41,316,087 \\
    \hline
    26 & 275,654,944 & 136,723,841 & \textbf{21,806,460} & 28,917,117 \\
    \hline
    27 & 294,354,320 & 13,3290,131 & 25,792,146 & \textbf{22,497,325} \\
    \hline
    28 & 297,756,076 & 195,971,759 & \textbf{34,602,990} & 39,975,903 \\
    \hline
    29 & 317,748,541 & 78,399,969 & 13,360,829 & \textbf{6,625,366} \\
    \hline
      \end{tabular}
 \caption{Average Cumulative Costs in \$ obtained over $100$ replications, horizon of $T=240$ months for Items 20-29.}
    \label{tab:costs_20_29}
    % }
    % \hfill
    % \parbox{.45\linewidth}{
\end{table}

\begin{table}[h!]
    \centering
      \begin{tabular}{crrrr}
       \hline
       ID & \multicolumn{1}{c}{MinMax} & \multicolumn{1}{c}{Oracle} & \multicolumn{1}{c}{PPO-D} & \multicolumn{1}{c}{PPO-C} \\
    \hline
    30 & 340,850,617 & 118,356,448 & 8,428,058 & \textbf{5,674,420} \\
    \hline
    31 & 346,677,835 & 165,513,863 & \textbf{38,211,495} & 50,257,711 \\
    \hline
    32 & 372,022,376 & 140,849,611 & 20,580,162 & \textbf{13,426,053} \\
    \hline
    33 & 379,189,787 & 176,014,600 & 17,857,314 & \textbf{12,033,561} \\
    \hline
    34 & 379,417,845 & 187,467,258 & 20,350,345 & \textbf{18,721,088} \\
    \hline
    35 & 429,201,653 & 125,815,263 & 15,920,620 & \textbf{10,148,258} \\
    \hline
    36 & 450,249,676 & 197,138,780 & \textbf{20,805,554} & 21,535,008 \\
    \hline
    37 & 467,942,296 & 221,381,388 & \textbf{40,233,046} & 51,880,344 \\
    \hline
    38 & 517,401,650 & 217,431,556 & \textbf{25,188,489} & 35,934,573 \\
    \hline
    39 & 529,054,715 & 113,253,016 & 35,899,378 & \textbf{19,663,470} \\
    \hline
        \end{tabular}
    \caption{Average Cumulative Costs in \$ obtained over $100$ replications, horizon of $T=240$ months for Items 30-39.}
    \label{tab:costs_30_39}
    % }
\end{table}

\begin{table}[h!]
    \centering
    \begin{tabular}{crrrr}
       \hline
       ID & \multicolumn{1}{c}{MinMax} & \multicolumn{1}{c}{Oracle} & \multicolumn{1}{c}{PPO-D} & \multicolumn{1}{c}{PPO-C} \\
    \hline
    40 & 548,632,612 & 274,992,151 & 19,614,638 & \textbf{14,265,377} \\
    \hline
    41 & 614,478,713 & 351,832,711 & \textbf{91,357,970} & 158,742,424 \\
    \hline
    42 & 802,247,508 & 318,672,823 & 28,692,777 & \textbf{22,605,438} \\
    \hline
    43 & 826,810,789 & 277,103,747 & \textbf{13,143,955} & 18,262,774 \\
    \hline
    44 & 897,278,509 & 593,855,529 & \textbf{278,916,815} & 434,302,595 \\
    \hline
    45 & 1205,603,289 & 856,462,764 & 193,219,445 & \textbf{184,847,412} \\
    \hline
    46 & 1463,640,790 & 1196,188,961 & \textbf{1013,385,300} & 1137,694,029 \\
    \hline
    47 & 7013,730,126 & 5018,246,734 & \textbf{4692,465,941} & 5153,720,918 \\
    \hline
    48 & 7033,366,188 & 4830,858,675 & \textbf{4348,154,722} & 4928,213,014 \\
    \hline
    49 & 10354,442,106 & 9275,492,362 & \textbf{9137,883,131} & 9309,276,387 \\
    \hline
    \end{tabular}
    \caption{Average Cumulative Costs in \$ obtained over $100$ replications, horizon of $T=240$ months for Items 40-49.}
    \label{tab:costs_40_49}
\end{table}
%%%%%%%%%%%%%%%%%%%%%%%%%%%%%%%%%%%%%%%%%%%%%%%%%%%%%%%%%%%%%%%%%%%%%%%%%%%%%%%%%%
%%%%%%%%%%%%%%%%%%%%%%%%%%%%%%%%%%%%%%%%%%%%%%%%%%%%%%%%%%%%%%%%%%%%%%%%%%%%%%%%%
\clearpage
\newpage
\subsection{Barplots results}
Similarly to Section \ref{subsec:num_results_single}, the different barplots below present the full results of cumulative costs for the numerical experiments with storage space constraints per item. The $50$ items are split in $2$ groups according to the median of lead-time in order to train only $2$ average RL agent. Such agents are then tested on all the items in the corresponding group.

\begin{figure}[h!]
\centering
\begin{minipage}{.48\textwidth}
  \centering
  \includegraphics[width=\linewidth]{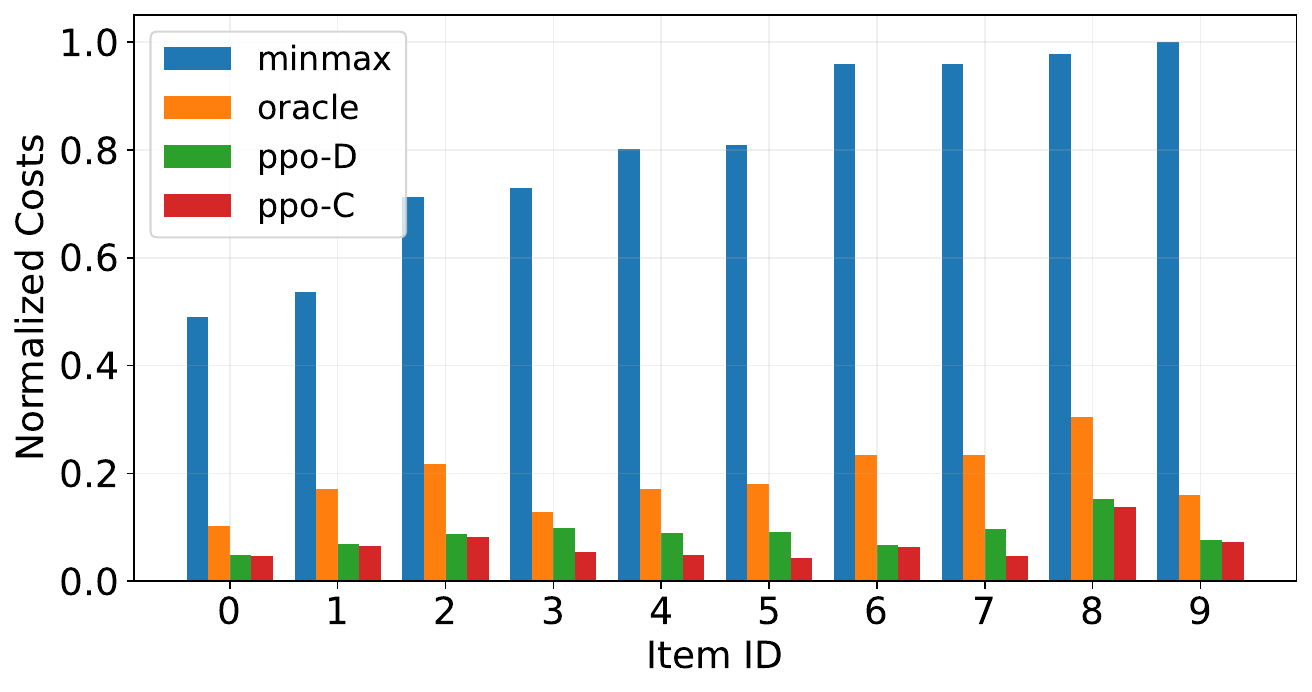}
  \caption{Average Cumulative Costs in \$ obtained over $100$ replications, horizon of $T=240$ months for Items 0-9.}
  \label{fig:barplot_cost_0_9}
\end{minipage}%
\hfill
\begin{minipage}{.48\textwidth}
  \centering
  \includegraphics[width=\linewidth]{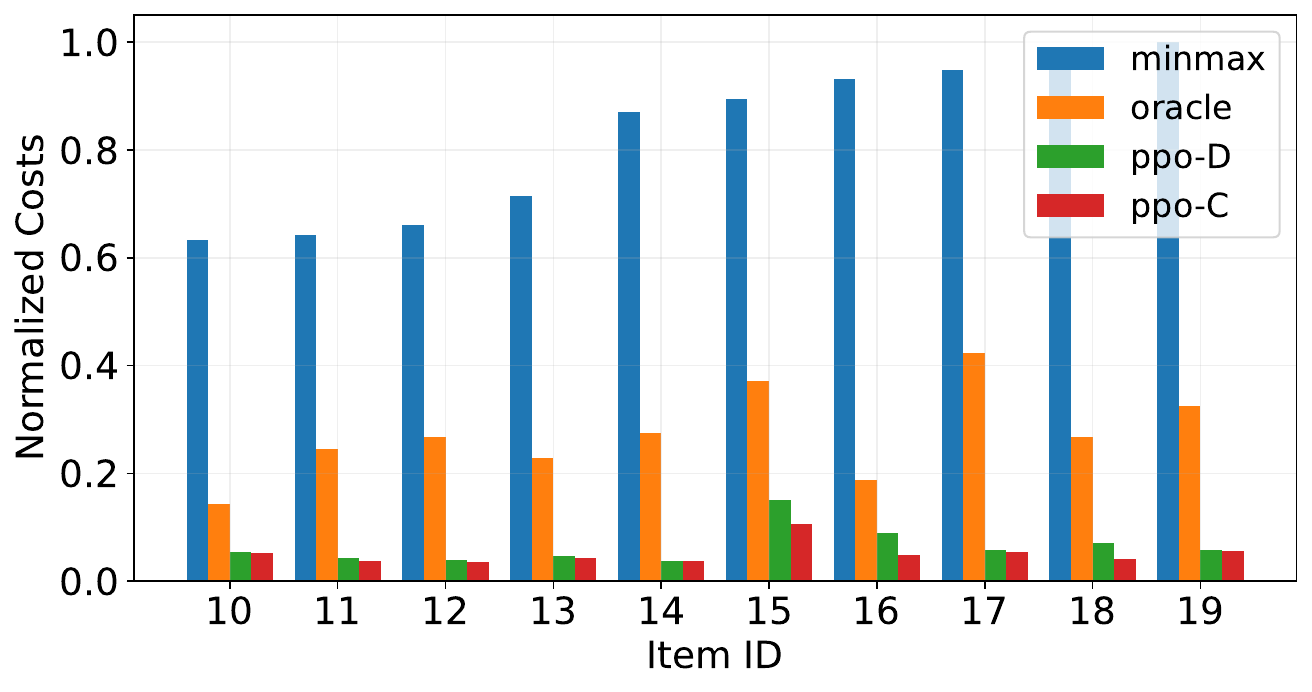}
  \caption{Average Cumulative Costs in \$ obtained over $100$ replications, horizon of $T=240$ months for Items 10-19.}
  \label{fig:barplot_cost_10_19}
\end{minipage}
\end{figure}

\begin{figure}[h!]
\centering
\begin{minipage}{.48\textwidth}
  \centering
  \includegraphics[width=\linewidth]{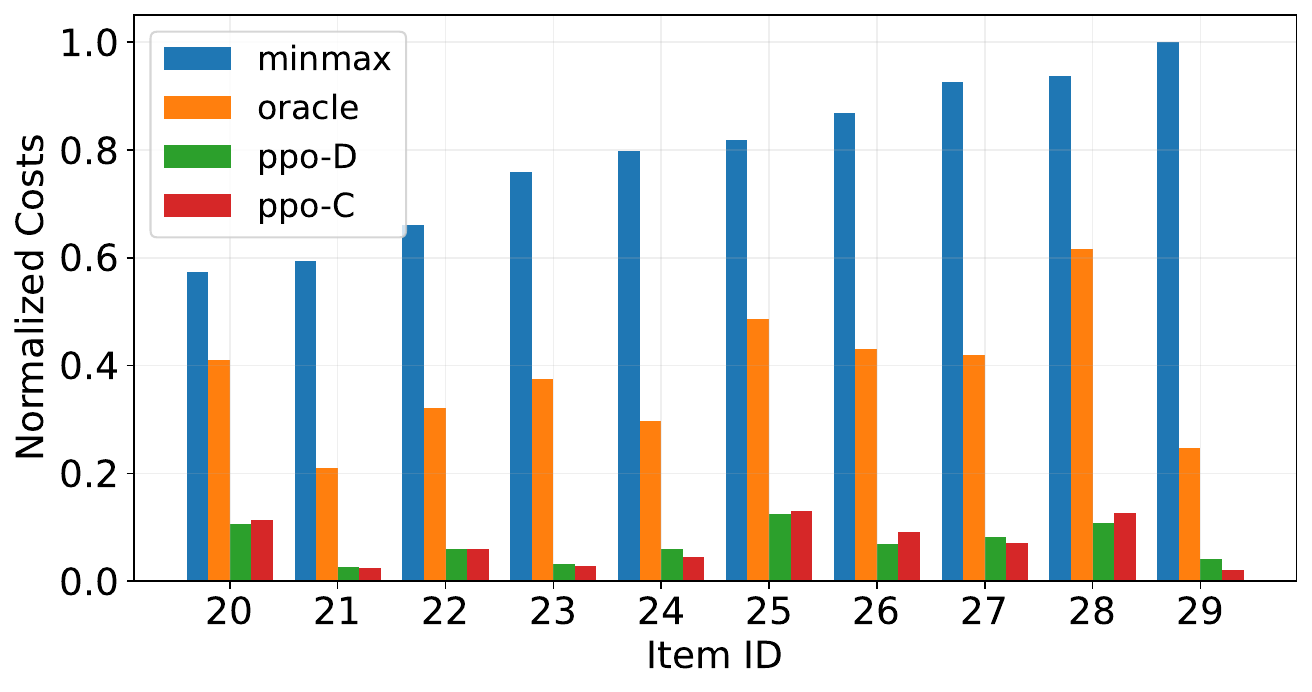}
  \caption{Average Cumulative Costs in \$ obtained over $100$ replications, horizon of $T=240$ months for Items 20-29.}
  \label{fig:barplot_cost_20_29}
\end{minipage}%
\hfill
\begin{minipage}{.48\textwidth}
  \centering
  \includegraphics[width=\linewidth]{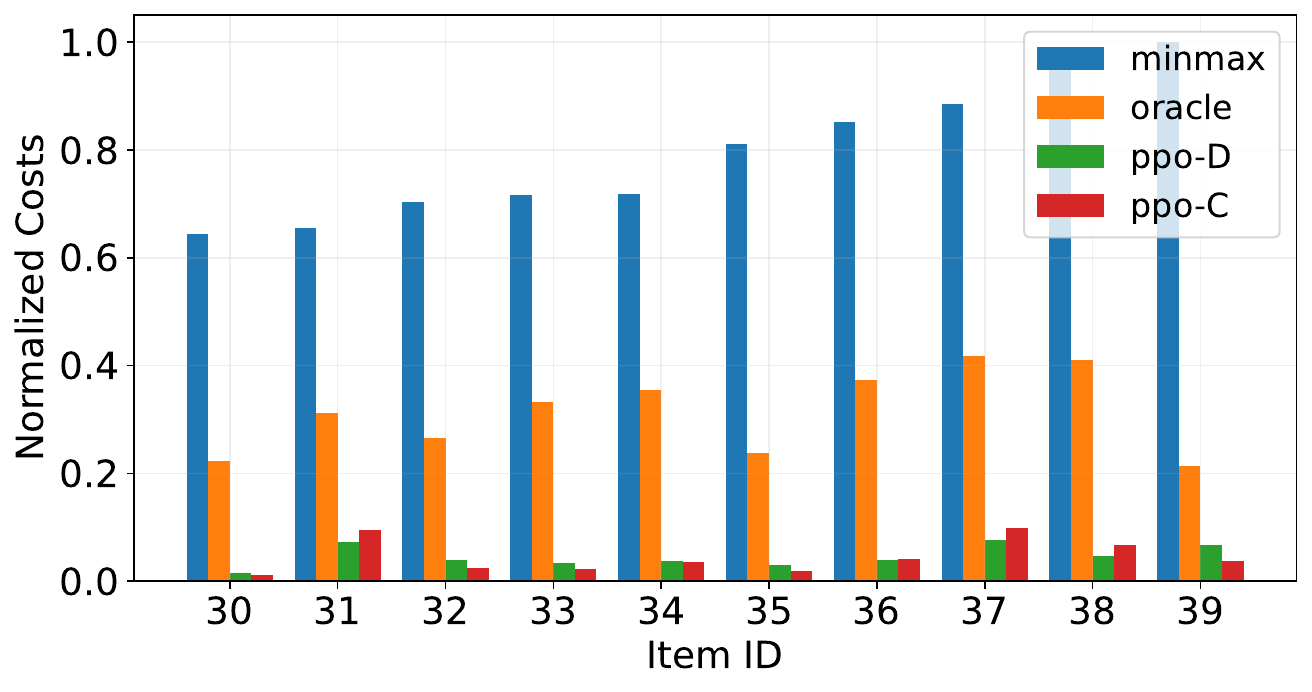}
  \caption{Average Cumulative Costs in \$ obtained over $100$ replications, horizon of $T=240$ months for Items 30-39.}
  \label{fig:barplot_cost_30_39}
\end{minipage}
\end{figure}

\begin{figure}[h!]
    \centering
    \includegraphics[width=\linewidth/2]{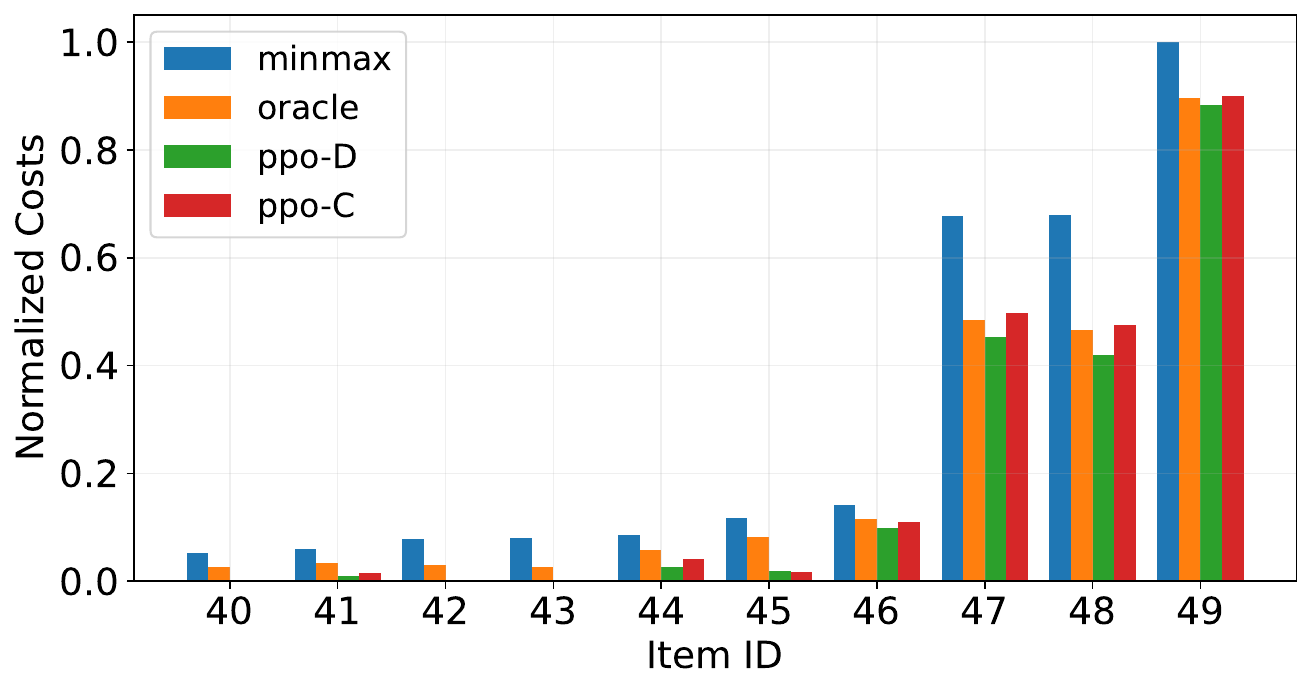}
    \caption{Average Cumulative Costs in \$ obtained over $100$ replications, horizon of $T=240$ months for Items 40-49.}
    \label{fig:barplot_cost_40_49}
\end{figure}

\newpage
\section{Average Item Shortages}
\label{app:shortages_single}

\subsection{Numerical results}
\label{subsec:num_results_shortage}

The different Tables below present the full results of item shortages for the numerical experiments with storage space constraints per item. The $50$ items are split in $2$ groups according to the median of lead-time in order to train only $2$ average RL agent. Such agents are then tested on all the items in the corresponding group.

\begin{table}[h!]
    \centering
    \parbox{.45\linewidth}{
    \begin{tabular}{ccccc}
    \hline
       ID & \multicolumn{1}{c}{MinMax} & \multicolumn{1}{c}{Oracle} & \multicolumn{1}{c}{PPO-D} & \multicolumn{1}{c}{PPO-C} \\
   \hline
 0 & 39 & 10 & 0 & 0 \\
\hline
 1 & 41 & 11 & 0 & 0 \\
\hline
 2 & 42 & 13 & 0 & 0 \\
\hline
3 & 51 & 8 & 0 & 0 \\
\hline
4 & 49 & 10 & 0 & 0 \\
\hline
5 & 43 & 10 & 0 & 0 \\
\hline
6 & 47 & 11 & 0 & 0 \\
\hline
7 & 57 & 12 & 0 & 0 \\
\hline
8 & 38 & 10 & 0 & 0 \\
\hline
9 & 59 & 9 & 0 & 0 \\
\hline
 \end{tabular}
    \caption{Average Item shortages obtained over $100$ replications, horizon of $T=240$ months for Items 0-9.}
    }
    \hfill
\parbox{.45\linewidth}{
    \begin{tabular}{ccccc}
    \hline
       ID & \multicolumn{1}{c}{MinMax} & \multicolumn{1}{c}{Oracle} & \multicolumn{1}{c}{PPO-D} & \multicolumn{1}{c}{PPO-C} \\
    \hline
    10 & 44 & 11 & 0 & 0 \\
    \hline
    11 & 69 & 23 & 0 & 0 \\
    \hline
    12 & 66 & 26 & 0 & 0 \\
    \hline
    13 & 76 & 25 & 0 & 0 \\
    \hline
    14 & 119 & 37 & 0 & 0 \\
    \hline
    15 & 37 & 14 & 0 & 0 \\
    \hline
    16 & 65 & 13 & 0 & 0 \\
    \hline
    17 & 132 & 58 & 2 & 2 \\
    \hline
    18 & 84 & 22 & 0 & 0 \\
    \hline
    19 & 71 & 26 & 0 & 0 \\
    \hline
\end{tabular}
    \caption{Average Item shortages obtained over $100$ replications, horizon of $T=240$ months for Items 10-19.}
    \label{tab:products}
    }
\end{table}

\begin{table}[h!]
    \centering
    \parbox{.45\linewidth}{
    \begin{tabular}{ccccc}
    \hline
       ID & \multicolumn{1}{c}{MinMax} & \multicolumn{1}{c}{Oracle} & \multicolumn{1}{c}{PPO-D} & \multicolumn{1}{c}{PPO-C} \\
    \hline
    20 & 125 & 84 & 16 & 17 \\
    \hline
    21 & 119 & 41 & 0 & 0 \\
    \hline
    22 & 108 & 52 & 3 & 5 \\
    \hline
    23 & 132 & 69 & 1 & 1 \\
    \hline
    24 & 95 & 29 & 0 & 0 \\
    \hline
    25 & 155 & 88 & 16 & 16 \\
    \hline
    26 & 200 & 89 & 7 & 14 \\
    \hline
    27 & 109 & 46 & 2 & 4 \\
    \hline
    28 & 182 & 114 & 15 & 18 \\
    \hline
    29 & 109 & 26 & 0 & 0 \\
    \hline
     \end{tabular}
       \caption{Average Item shortages obtained over $100$ replications, horizon of $T=240$ months for Items 20-29.}
        }
    \hfill
\parbox{.45\linewidth}{
    \begin{tabular}{ccccccc}
    \hline
        ID & \multicolumn{1}{c}{MinMax} & \multicolumn{1}{c}{Oracle} & \multicolumn{1}{c}{PPO-D} & \multicolumn{1}{c}{PPO-C} \\
 \hline
30 & 219 & 69 & 0 & 0 \\
\hline
31 & 209 & 94 & 14 & 22 \\
\hline
32 & 140 & 48 & 0 & 0 \\
\hline
33 & 234 & 110 & 4 & 2 \\
\hline
34 & 196 & 95 & 6 & 5 \\
\hline
35 & 197 & 49 & 0 & 0 \\
\hline
36 & 155 & 66 & 3 & 3 \\
\hline
37 & 193 & 89 & 9 & 16 \\
\hline
38 & 220 & 86 & 5 & 11 \\
\hline
 39& 90 & 18 & 0 & 0 \\
\hline
\end{tabular}
    \caption{Average Item shortages obtained over $100$ replications, horizon of $T=240$ months for Items 30-39.}
    \label{tab:products}
    }
\end{table}

\begin{table}[h!]
    \centering
    \begin{tabular}{ccccc}
    \hline
       ID & \multicolumn{1}{c}{MinMax} & \multicolumn{1}{c}{Oracle} & \multicolumn{1}{c}{PPO-D} & \multicolumn{1}{c}{PPO-C} \\
   \hline
40 & 209 & 102 & 2 & 1 \\
\hline
41 & 305 & 172 & 40 & 75 \\
\hline
42 & 240 & 95 & 3 & 3 \\
\hline
43 & 472 & 141 & 1 & 5 \\
\hline
44 & 575 & 360 & 167 & 264 \\
\hline
45 & 276 & 186 & 35 & 33 \\
\hline
46 & 1230 & 986 & 837 & 945 \\
\hline
47 & 3814 & 2695 & 2520 & 2770 \\
\hline
48 & 2959 & 1989 & 1778 & 2030 \\
\hline
49 & 3453 & 3066 & 3025 & 3081 \\
\hline
    \end{tabular}
    \caption{Average Item shortages obtained over $100$ replications, horizon of $T=240$ months for Items 40-49.}
    \label{tab:item_shortages}
\end{table}
%%%%%%%%%%%%%%%%%%%%%%%%%%%%%%%%%%%%%%%%%%%%%%%%%%%%%%%%%%%%%%%%%%%%%%%%%
\newpage
\subsection{Barplots results}

Similarly to Section \ref{subsec:num_results_shortage}, the different barplots below present the full results of item shortages for the numerical experiments with storage space constraints per item. The $50$ items are split in $2$ groups according to the median of lead-time in order to train only $2$ average RL agent. Such agents are then tested on all the items in the corresponding group.

\begin{figure}[h!]
\centering
\begin{minipage}{.48\textwidth}
  \centering
  \includegraphics[width=\linewidth]{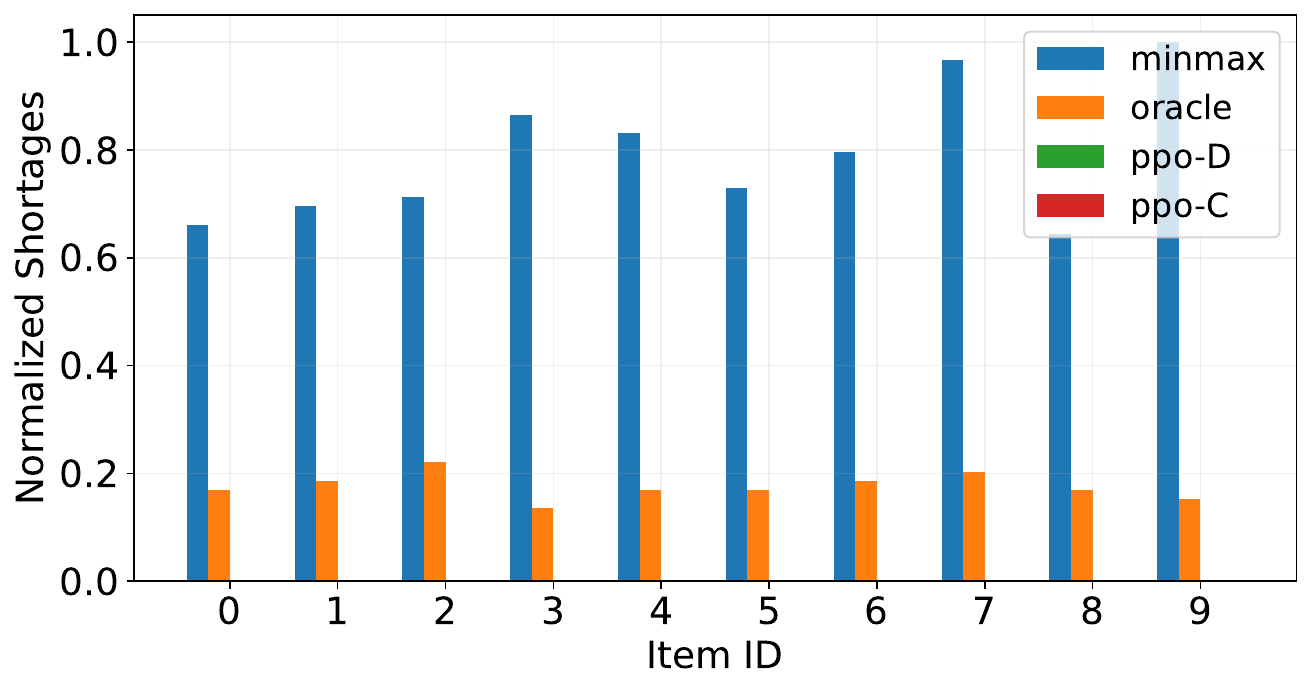}
  \caption{Average Item shortages obtained over $100$ replications, horizon of $T=240$ months for Items 0-9.}
  \label{fig:barplot_shortage_0_9}
\end{minipage}%
\hfill
\begin{minipage}{.48\textwidth}
  \centering
  \includegraphics[width=\linewidth]{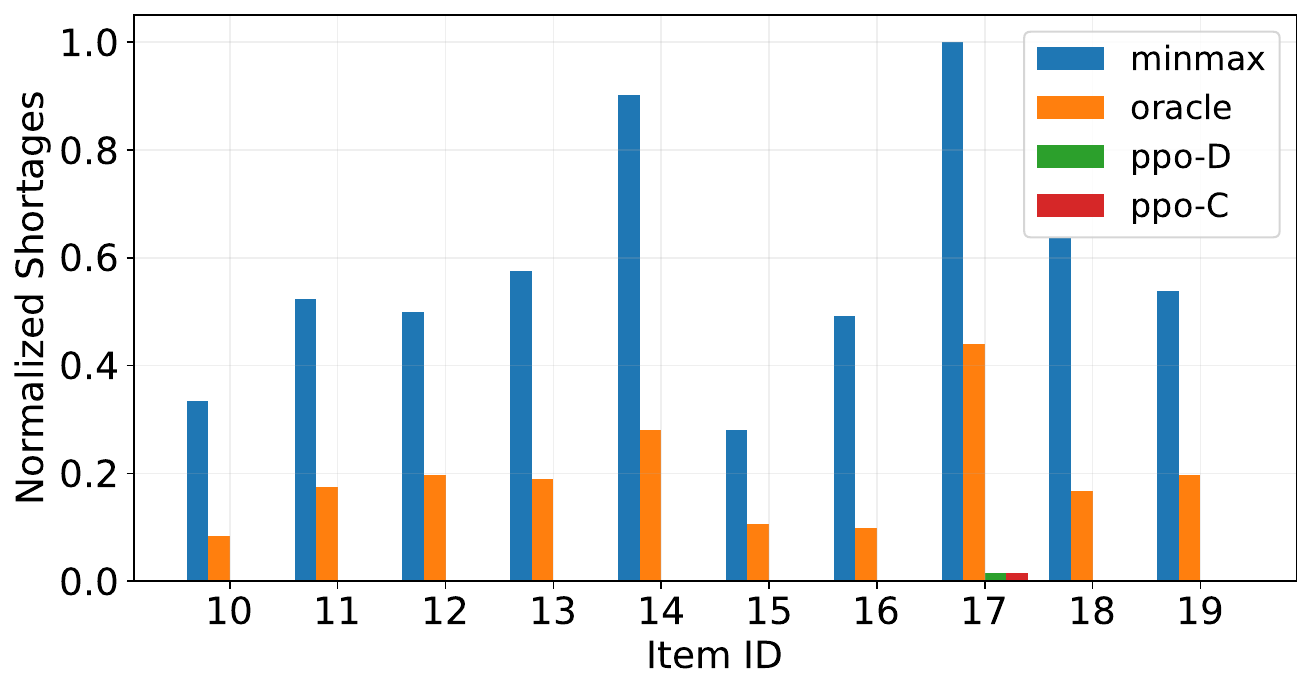}
  \caption{Average Item shortages obtained over $100$ replications, horizon of $T=240$ months for Items 10-19.}
  \label{fig:barplot_shortage_10_19}
\end{minipage}
\end{figure}

\begin{figure}[h!]
\centering
\begin{minipage}{.48\textwidth}
  \centering
  \includegraphics[width=\linewidth]{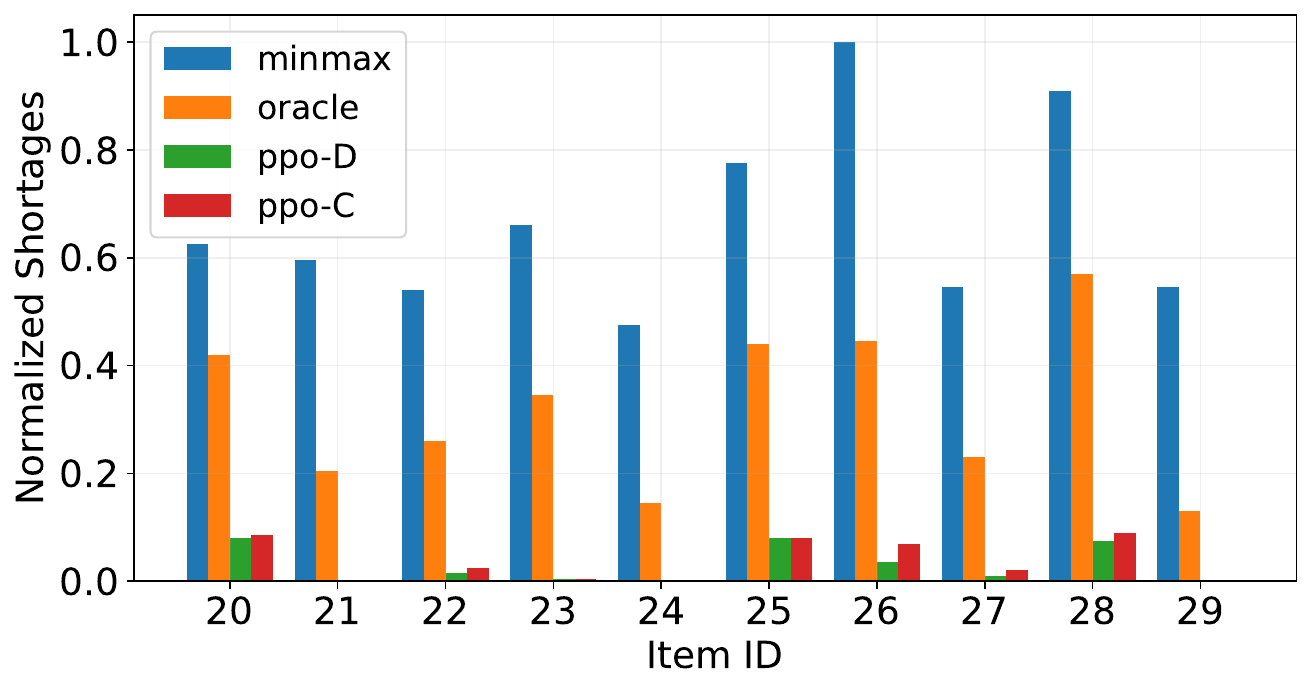}
  \caption{Average Item shortages obtained over $100$ replications, horizon of $T=240$ months for Items 20-29.}
  \label{fig:barplot_shortage_20_29}
\end{minipage}%
\hfill
\begin{minipage}{.48\textwidth}
  \centering
  \includegraphics[width=\linewidth]{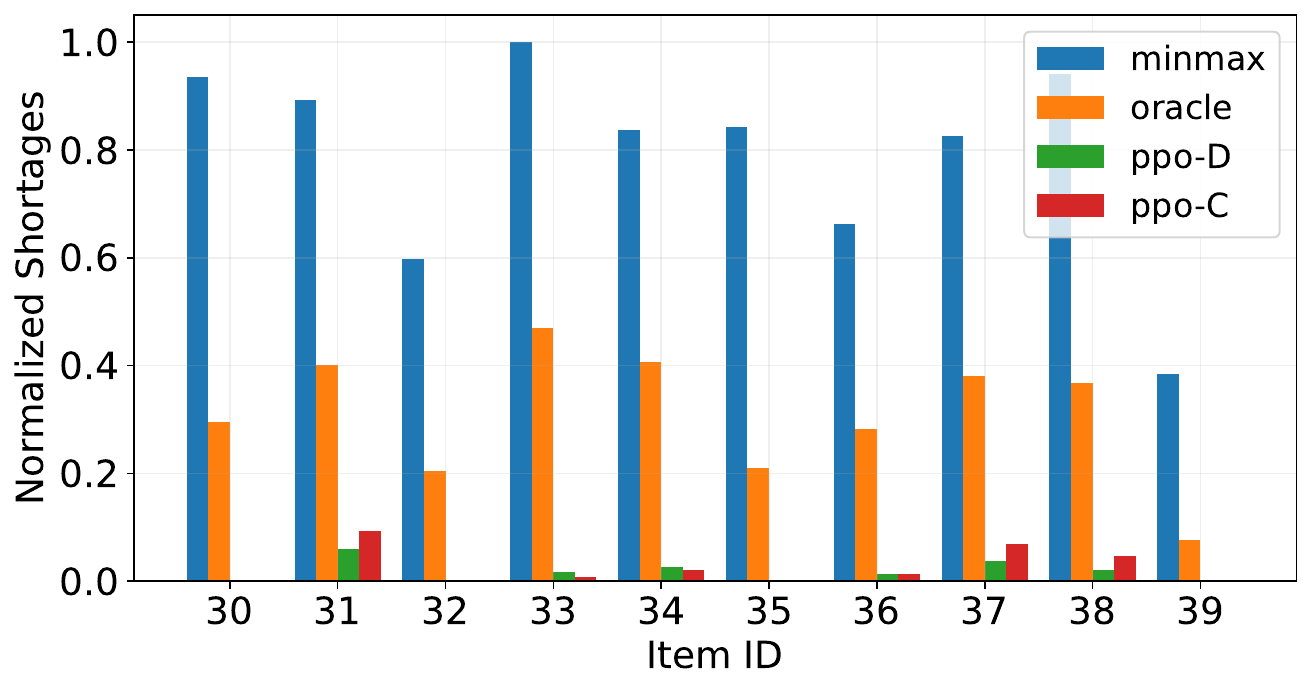}
  \caption{Average Item shortages obtained over $100$ replications, horizon of $T=240$ months for Items 30-39.}
  \label{fig:barplot_shortage_30_39}
\end{minipage}
\end{figure}

\begin{figure}[h!]
    \centering
    \includegraphics[width=\linewidth/2]{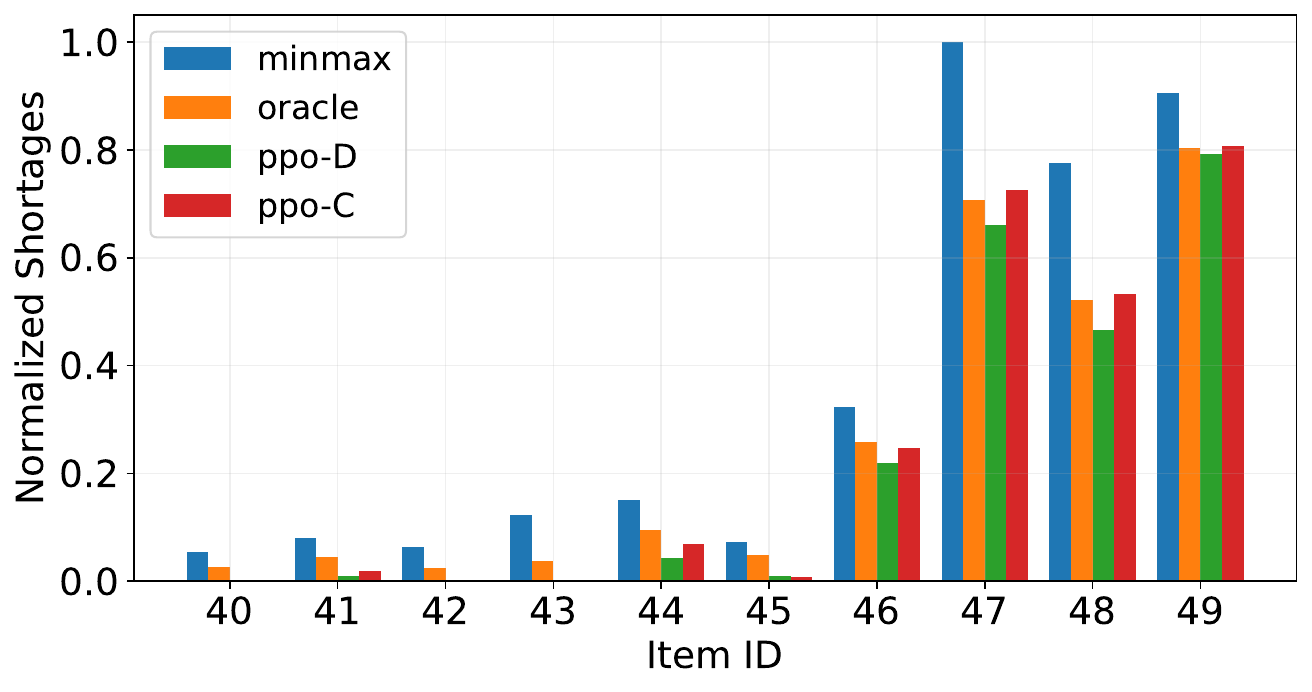}
    \caption{Average Item shortages obtained over $100$ replications, horizon of $T=240$ months for Items 40-49.}
    \label{fig:barplot_shortage_40_49}
\end{figure}

\newpage
\section{Average Cumulative Costs and Item Shortages for Multi-Agent cases}
\label{app:costs_multi}

\subsection{Cluster $\mathcal{N}_1$}

\begin{table}[h!]
    \centering
    % \parbox{.45\linewidth}{
    \centering
    \begin{tabular}{crrrr}
       \hline
       ID & \multicolumn{1}{c}{MinMax} & \multicolumn{1}{c}{Oracle} & \multicolumn{1}{c}{IPPO-C} \\
    \hline
    $\mathcal{N}_1$-0 & 56,413,703 & 47,751,643 & \textbf{21,747,138} \\
    \hline
    $\mathcal{N}_1$-1 & 41,218,939 & 22,017,147 & \textbf{5,860,091} \\
    \hline
    $\mathcal{N}_1$-2 & 28,734,942 & 10,052,437 & \textbf{9,674,691} \\
    \hline
    $\mathcal{N}_1$-3 & 69,445,374 & 49,567,318 & \textbf{11,468,600} \\
    \hline
    $\mathcal{N}_1$-4 & 50,283,017 & 18,591,579 & \textbf{11,492,810} \\
    \hline
    Average & 49,219,195 & 29,596,024 & \textbf{12,048,666} \\
    \hline
     \end{tabular}
    \caption{Average Cumulative Costs in \$ obtained over $100$ replications, horizon of $T=240$ months for cluster $\mathcal{N}_1$}
    \label{tab:costs_cluster_N1}
    \end{table}
\begin{table}[h!]
    % }
    % \hfill
    % \parbox{.45\linewidth}{
    \centering
     \begin{tabular}{crrrr}
       \hline
       ID & \multicolumn{1}{c}{MinMax} & \multicolumn{1}{c}{Oracle} & \multicolumn{1}{c}{IPPO-C} \\
    \hline
    $\mathcal{N}_1$-0 & 27 & 25 & \textbf{0} \\
    \hline
    $\mathcal{N}_1$-1 & 18 & 10 & \textbf{1} \\
    \hline
    $\mathcal{N}_1$-2 & 11 & \textbf{4} & 6 \\
        \hline
    $\mathcal{N}_1$-3 & 22 & 21 & \textbf{5} \\
        \hline
    $\mathcal{N}_1$-4 & 14 & 5 & \textbf{3} \\
    \hline
    Average & 18 & 9 & \textbf{3} \\
    \hline
     \end{tabular}
    \caption{Average Item Shortages obtained over $100$ replications, horizon of $T=240$ months for cluster $\mathcal{N}_1$}
    \label{tab:shortages_cluster_N1}
    % }
    \end{table}
    
\begin{figure}[h]
\hfill
\includegraphics[width=6cm]{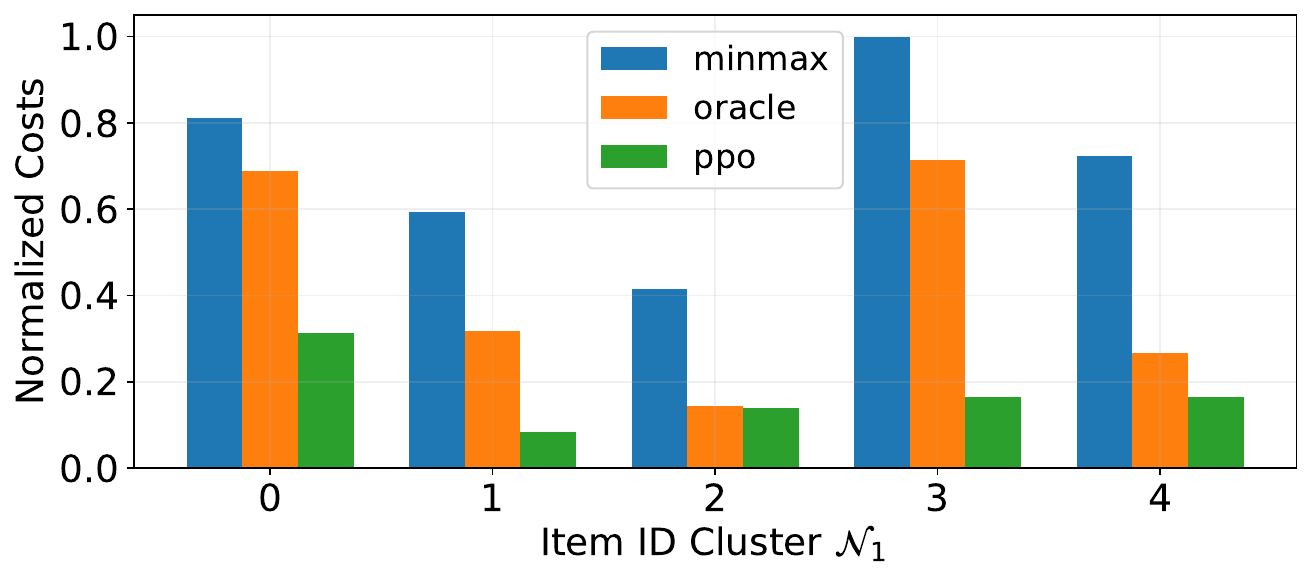}
\hfill
\includegraphics[width=6cm]{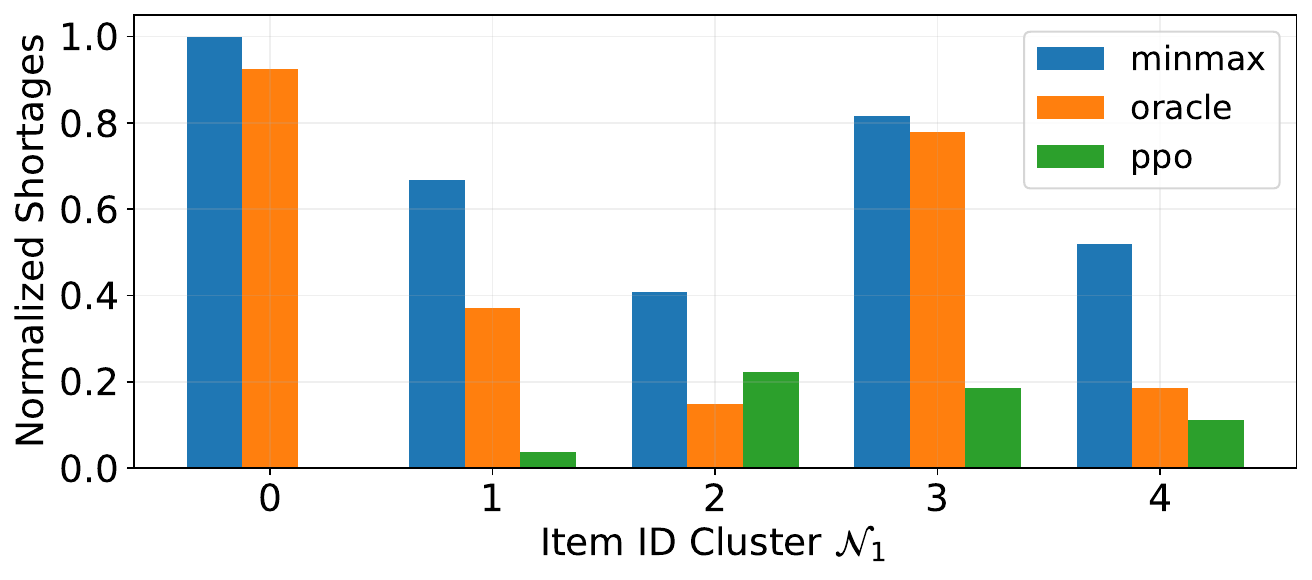}
\hfill
\caption{Average Cumulative Costs in \$ and Item Shortages obtained over $100$ replications, horizon of $T=240$ months for cluster $\mathcal{N}_1$}
\end{figure}

\subsection{Cluster $\mathcal{N}_2$}
\begin{table}[h!]
    \centering
    % \parbox{.45\linewidth}{
    \begin{tabular}{crrrr}
       \hline
       ID & \multicolumn{1}{c}{MinMax} & \multicolumn{1}{c}{Oracle} & \multicolumn{1}{c}{IPPO-C} \\
    \hline
    $\mathcal{N}_2$-0 & 61,746,686 & 37,154,240 & \textbf{9,576,664} \\
    \hline
    $\mathcal{N}_2$-1 & 43,195,470 & 19,817,180 & \textbf{4,755,388} \\
    \hline
    $\mathcal{N}_2$-2 & 31,719,337 & 8,849,048 & \textbf{7,232,925} \\
    \hline
    $\mathcal{N}_2$-3 & 49,873,610 & 38,995,232 & \textbf{7,999,544} \\
    \hline
    $\mathcal{N}_2$-4 & 46,383,922 & 17,470,651 & \textbf{8,097,309} \\
    \hline
    $\mathcal{N}_2$-5 & 90,361,502 & 54,257,529 & \textbf{6,974,378} \\
    \hline
    $\mathcal{N}_2$-6 & 20,349,092 & 8,573,083 & \textbf{4,180,512} \\
    \hline
    $\mathcal{N}_2$-7 & 27,434,488 & 9,091,921 & \textbf{9,855,567} \\
    \hline
    $\mathcal{N}_2$-8 & 33,681,010 & 11,145,783 & \textbf{3,862,856} \\
    \hline
    $\mathcal{N}_2$-9 & 146,735,155 & 105,116,776 & \textbf{16,516,513} \\
    \hline
    Average & 55,148,027 & 31,047,144 &  \textbf{7,905,165} \\
    \hline
     \end{tabular}
    \caption{Average Cumulative Costs in \$ obtained over $100$ replications, horizon of $T=240$ months for cluster $\mathcal{N}_2$}
    \label{tab:costs_cluster_N2}
    % }
    % \hfill
    % \parbox{.45\linewidth}{
    \end{table}
\begin{table}[h!]
\centering
     \begin{tabular}{crrrr}
       \hline
       ID & \multicolumn{1}{c}{MinMax} & \multicolumn{1}{c}{Oracle} & \multicolumn{1}{c}{IPPO-C} \\
    \hline
    $\mathcal{N}_2$-0 & 33 & 18 & \textbf{1} \\
    \hline
    $\mathcal{N}_2$-1 & 18 & 8 & \textbf{0} \\
    \hline
    $\mathcal{N}_2$-2  & 14 & 2 & \textbf{0} \\
        \hline
    $\mathcal{N}_2$-3  & 16 & 14 & \textbf{0} \\
        \hline
    $\mathcal{N}_2$-4 & 13 & 4 & \textbf{0} \\
        \hline
    $\mathcal{N}_2$-5 & 27 & 16 & \textbf{0} \\
        \hline
    $\mathcal{N}_2$-6 & 11 & 8 & \textbf{0} \\
        \hline
    $\mathcal{N}_2$-7 & 11 & 2 & \textbf{0} \\
        \hline
    $\mathcal{N}_2$-8 & 14 & 5 & \textbf{0} \\
    \hline
    $\mathcal{N}_2$-9 & 20 & 13 & \textbf{0} \\
    \hline
    Average & 18 & 9 & \textbf{0} \\
    \hline
     \end{tabular}
    \caption{Average Item Shortages obtained over $100$ replications, horizon of $T=240$ months for cluster $\mathcal{N}_2$}
    \label{tab:shortages_cluster_N2}
    \end{table}

\begin{figure}[h!]
\hfill
\includegraphics[width=6cm]{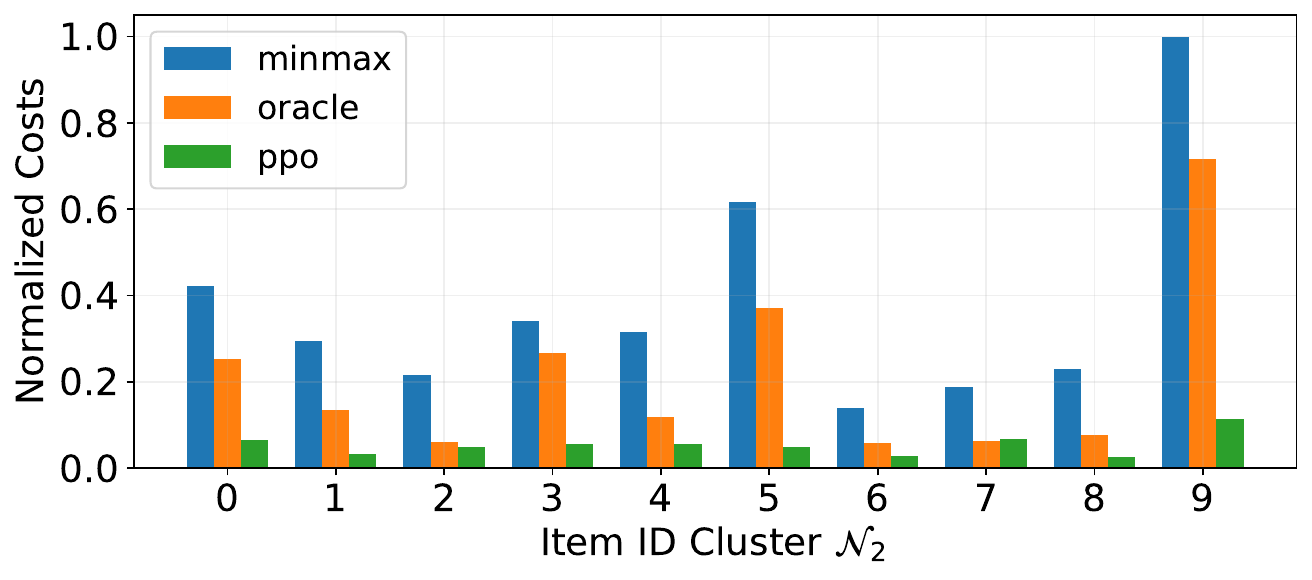}
\hfill
\includegraphics[width=6cm]{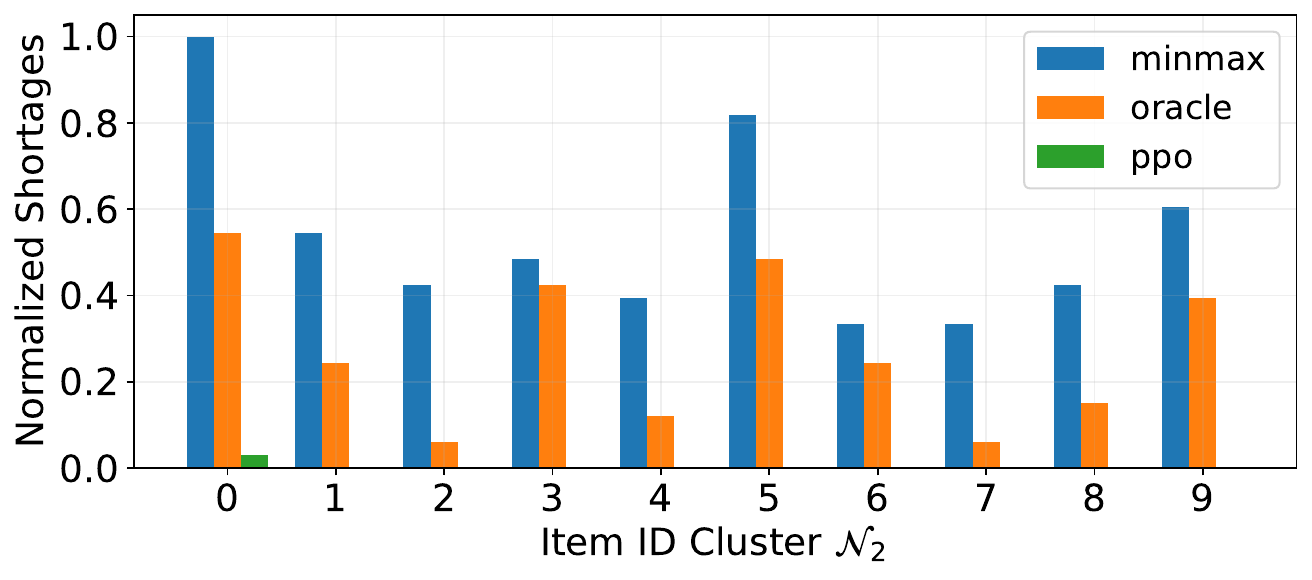}
\hfill
\caption{Average Cumulative Costs in \$ and Item Shortages obtained over $100$ replications, horizon of $T=240$ months for cluster $\mathcal{N}_2$}
\end{figure}

\newpage
\subsection{Cluster $\mathcal{N}_3$}

\begin{table}[h!]
    \centering
    % \parbox{.45\linewidth}{
    \begin{tabular}{crrrr}
       \hline
       ID & \multicolumn{1}{c}{MinMax} & \multicolumn{1}{c}{Oracle} & \multicolumn{1}{c}{IPPO-C} \\
    \hline
    $\mathcal{N}_3$-0 & 114,748,999 & 110,936,731 & \textbf{25,133,542} \\
    \hline
$\mathcal{N}_3$-1  & 76,876,589 & 25,248,863 & \textbf{13,012,784} \\
    \hline
$\mathcal{N}_3$-2  & 175,969,893 & 156,867,720 & \textbf{22,712,799} \\
    \hline
$\mathcal{N}_3$-3  & 61,746,686 & 38,243,787 & \textbf{11,097,544} \\
    \hline
$\mathcal{N}_3$-4  & 43,195,470 & 14,974,424 & \textbf{6,590,003} \\
    \hline
$\mathcal{N}_3$-5  & 31,719,337 & \textbf{9,423,716} & 12,452,378 \\
    \hline
$\mathcal{N}_3$-6  & 49,873,610 & 35,928,930 & \textbf{13,922,411} \\
    \hline
$\mathcal{N}_3$-7  & 46,383,922 & 20,756,443 & \textbf{7,299,909} \\
    \hline
$\mathcal{N}_3$-8  & 34,654,173 & \textbf{9,843,404} & 11,531,224 \\
    \hline
$\mathcal{N}_3$-9  & 90,361,502 & 55,657,119 & \textbf{13,186,238} \\
    \hline
$\mathcal{N}_3$-10   & 20,349,092 & 8,662,162 & \textbf{5,354,639} \\
    \hline
$\mathcal{N}_3$-11   & 41,008,234 & 12,947,589 & \textbf{10,195,819} \\
    \hline
$\mathcal{N}_3$-12   & 33,335,933 & 12,451,461 & \textbf{10,197,750} \\
    \hline
$\mathcal{N}_3$-13   & 27,434,488 & 8,713,145 & \textbf{8,518,849} \\
    \hline
$\mathcal{N}_3$-14  & 33,681,010 & 10,451,041 & \textbf{8,045,677} \\
    \hline
$\mathcal{N}_3$-15  & 106,676,800 & 56,572,852 & \textbf{11,317,268} \\
    \hline
$\mathcal{N}_3$-16  & 146,735,155 & 126,683,713 & \textbf{55,932,376} \\
    \hline
$\mathcal{N}_3$-17  & 46,569,841 & 24,907,569 & \textbf{14,623,596} \\
    \hline
$\mathcal{N}_3$-18  & 41,120,038 & \textbf{9,885,961} & 13,659,858 \\
    \hline
$\mathcal{N}_3$-19  & 43,945,213 & 10,883,913 & \textbf{10,006,678} \\
    \hline
    Average & 63,319,299 & 38,002,027 & \textbf{14,239,567} \\
    \hline
     \end{tabular}
    \label{tab:costs_cluster_N3}
    \caption{Average Cumulative Costs in \$ obtained over $100$ replications, horizon of $T=240$ months for cluster $\mathcal{N}_3$}
    
    % }
    % \hfill
    \end{table}
\begin{table}[h!]
\centering
    % \parbox{.45\linewidth}{
     \begin{tabular}{crrrr}
       \hline
       ID & \multicolumn{1}{c}{MinMax} & \multicolumn{1}{c}{Oracle} & \multicolumn{1}{c}{IPPO-C} \\
    \hline
    $\mathcal{N}_3$-0 & 39 & 33 & \textbf{7} \\
    \hline
    $\mathcal{N}_3$-1 & 19 & 4 & \textbf{0} \\
    \hline
    $\mathcal{N}_3$-2 & 97 & 63 & \textbf{0} \\
    \hline
    $\mathcal{N}_3$-3 & 33 & 19 & \textbf{1} \\
    \hline
    $\mathcal{N}_3$-4 & 18 & 5 & \textbf{0} \\
    \hline
    $\mathcal{N}_3$-5 & 14 & \textbf{4} & 6 \\
    \hline
    $\mathcal{N}_3$-6 & 16 & 14 & \textbf{4} \\
    \hline
    $\mathcal{N}_3$-7 & 13 & 5 & \textbf{0} \\
    \hline
    $\mathcal{N}_3$-8 & 15 & 2 & 2 \\
    \hline
    $\mathcal{N}_3$-9 & 27 & 17 & \textbf{4} \\
    \hline
    $\mathcal{N}_3$-10 & 11 & 6 & \textbf{1} \\
    \hline
    $\mathcal{N}_3$-11 & 9 & 0 & 0 \\
    \hline
    $\mathcal{N}_3$-12 & 13 & 3 & 3 \\
    \hline
    $\mathcal{N}_3$-13 & 11 & 3 & \textbf{0} \\
    \hline
    $\mathcal{N}_3$-14 & 14 & 4 & \textbf{2} \\
    \hline
    $\mathcal{N}_3$-15 & 55 & 22 & \textbf{0} \\
    \hline
    $\mathcal{N}_3$-16 & 20 & 18 & \textbf{2} \\
    \hline
    $\mathcal{N}_3$-17 & 19 & 7 & \textbf{7} \\
    \hline
    $\mathcal{N}_3$-18 & 14 & 2 & \textbf{0} \\
    \hline
    $\mathcal{N}_3$-19 & 16 & \textbf{2} & 3 \\
    \hline
    Average & 23 & 11 & \textbf{2} \\
    \hline
     \end{tabular}
    \label{tab:shortages_cluster_N3}
    \caption{Average Item Shortages obtained over $100$ replications, horizon of $T=240$ months for cluster $\mathcal{N}_3$}
    % }
    \end{table}
    
\begin{figure}[h!]
\hfill
\includegraphics[width=6.5cm]{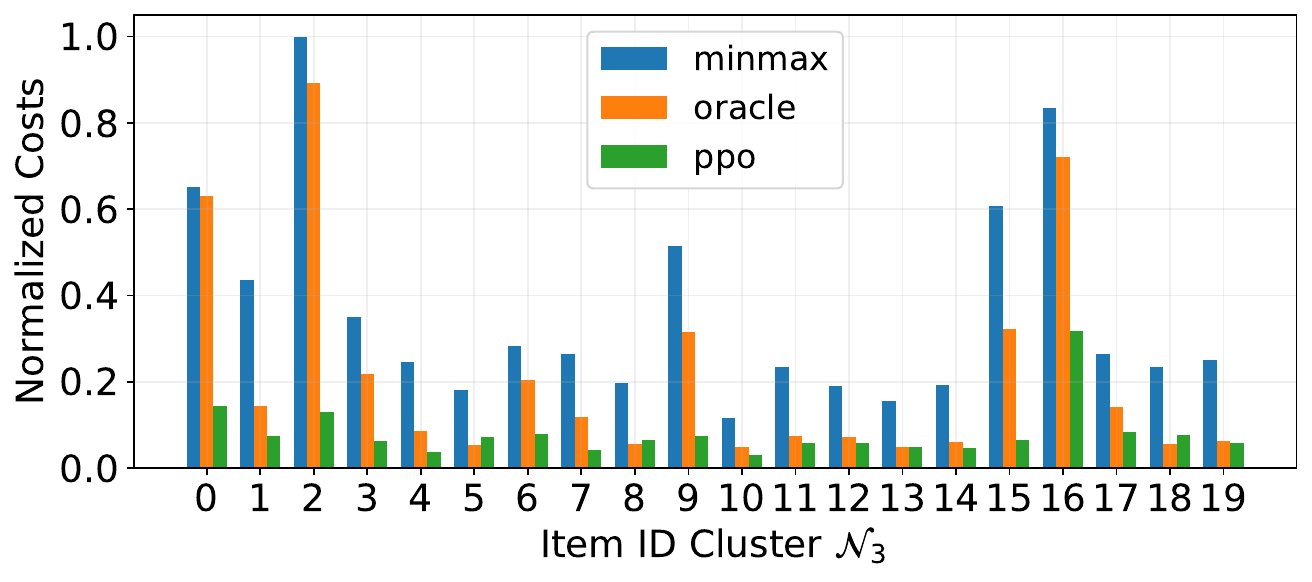}
\hfill
\includegraphics[width=6.5cm]{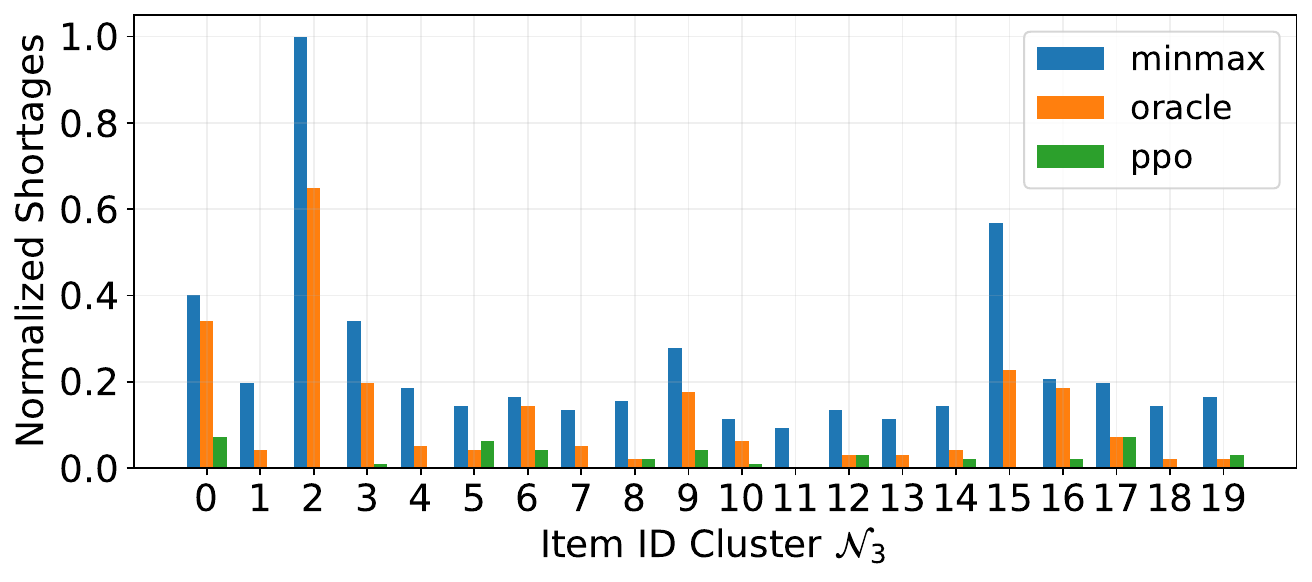}
\hfill
\caption{Average Cumulative Costs in \$ and Item Shortages obtained over $100$ replications, horizon of $T=240$ months for cluster $\mathcal{N}_3$}
\end{figure}

\end{document}